\documentclass[]{interact}

\usepackage{lineno}
\usepackage{epstopdf}% To incorporate .eps illustrations using PDFLaTeX, etc.
\usepackage[caption=false]{subfig}% Support for small, `sub' figures and tables
\usepackage{xcolor}
\usepackage{cancel}

\usepackage[numbers,sort&compress]{natbib}% Citation support using natbib.sty
\bibpunct[, ]{[}{]}{,}{n}{,}{,}% Citation support using natbib.sty
\theoremstyle{plain}% Theorem-like structures provided by amsthm.sty

\theoremstyle{definition}

\theoremstyle{remark}

\newcommand{\markupdraft}[2]{% {#1: {color|display} command}{#2: desired color or text}
    \ifthenelse{\equal{#1}{display}}{#2}{}%                 % display only in draft version
    \ifthenelse{\equal{#1}{color}}{\color{#2}}{}%           % colored only in draft (for \new command)
}
\renewcommand{\markupdraft}[2]{}  % remove all todo's, notes and coloring of changes
\newcommand{\nnotecolored}[3][]{\markupdraft{display}{{\color{#2}\noindent[#1: #3]}}}
\newcommand{\newcolored}[3][]{{\markupdraft{color}{#2}#3}%  % kept in the final print
    \ifthenelse{\equal{#1}{}}{}{\markupdraft{display}{{\color{yellow!70!black}[Pending Reviews: #1]}}}} 

\newcommand{\del}[2][]{{\markupdraft{display}{{\color{gray}[removed: "#2"[#1]]}}}} % (to be) removed
\newcommand{\new}[2][]{%\newcolored[#1]{blue!75!black}
{#2}}%  % kept in the final print
\newcommand{\nnew}[2][]{\newcolored[#1]{red}{#2}}%  % kept in the final print
    
\definecolor{frenchblue}{rgb}{0.0, 0.45, 0.73}
\definecolor{mediumblue}{rgb}{0.0, 0.0, 0.8}
\newcommand{\ddel}[2][]{{\markupdraft{display}{{\color{red}\cancel{#2}}}}} % (to be) removed

\newcommand{\todo}[2][]{\markupdraft{display}{{\color{red}\noindent++TODO: #2 {\color{yellow}(#1)}++}}}

\renewcommand{\del}[2][]{{}}
\renewcommand{\markupdraft}[2]{}
\newcommand{\niko}[1]{\protect\nnotecolored[Niko]{green!50!black}{~#1}}

\usepackage{hyperref}
\usepackage{hypcap}% it must be loaded after hyperref.
\usepackage{xspace}
\usepackage{listings}
\usepackage{xcolor}
\usepackage{framed}
\usepackage{mdframed}
\usepackage{threeparttable}

\newcommand{\COCO}{\href{https://github.com/numbbo/coco}{COCO}\xspace}
 
\newcommand{\R}{\ensuremath{\mathbb{R}}}
\newcommand{\ve}[1]{{\boldsymbol{#1}}}
\renewcommand{\x}{\ensuremath{\ve{x}}}
\newcommand{\finstance}{\ensuremath{f^j}}
\newcommand{\bbob}{{\ttfamily bbob}\xspace}
\newcommand{\bbobnoisy}{{\ttfamily bbob-noisy}\xspace}
\newcommand{\bbobbiobj}{{\ttfamily bbob-biobj}\xspace}
\newcommand{\bbobbiobjext}{{\ttfamily bbob-biobj-ext}\xspace}
\newcommand{\bboblargescale}{{\ttfamily bbob-largescale}\xspace}

\newcommand{\bbobmixint}{{\ttfamily bbob-mixint}\xspace}
\newcommand{\bbobbiobjmixint}{{\ttfamily bbob-biobj-mixint}\xspace}

\definecolor{codegreen}{rgb}{0,0.6,0}
\definecolor{codegray}{rgb}{0.5,0.5,0.5}
\definecolor{codepurple}{rgb}{0.58,0,0.82}
\definecolor{backcolour}{rgb}{0.95,0.95,0.92}
\lstdefinestyle{Pythonstyle}{
    backgroundcolor=\color{white},   
    commentstyle=\color{codegreen},
    keywordstyle=\color{magenta},
    numberstyle=\tiny\color{codegray},
    stringstyle=\color{codepurple},
    basicstyle=\ttfamily\scriptsize,
    breakatwhitespace=false,         
    breaklines=true,                 
    captionpos=b,                    
    keepspaces=true,                 
    numbers=none,                    
    numbersep=5pt,                  
    showspaces=false,                
    showstringspaces=false,
    showtabs=false,                  
    tabsize=2
}

\begin{document}
\pagestyle{plain}

%\articletype{ARTICLE TEMPLATE}% Specify the article type or omit as appropriate

\title{{COCO}: {A} Platform for Comparing Continuous Optimizers in a Black-Box Setting}

\author{
\name{Nikolaus Hansen\textsuperscript{a,b},
Anne Auger\textsuperscript{a,b},
\new{Raymond Ros\textsuperscript{c}},
Olaf Mersmann\textsuperscript{d},
Tea Tu\v{s}ar\textsuperscript{e},
and Dimo Brockhoff\textsuperscript{a,b}}
\affil{%
\textsuperscript{a}Inria, Randopt team, Palaiseau, France;
\textsuperscript{b}CMAP, CNRS, Ecole Polytechnique, Institut Polytechnique de Paris, 
Palaiseau, France;
\textsuperscript{c}\new{Universit\'{e}\ Paris-Sud, LRI, UMR 8623, Inria Saclay, France;}
%\textsuperscript{a}Inria, Randopt team, France;
%\textsuperscript{b}CMAP, CNRS, Ecole Polytechnique, Institut Polytechnique de Paris, France;
%\textsuperscript{c}Univ.\ Paris-Sud, LRI, Inria Tao team, France;
\textsuperscript{d}TU Dortmund University, Chair of Computational Statistics, Germany;
\textsuperscript{e}Jo\v{z}ef Stefan Institute, Ljubljana, Slovenia
}
}

\maketitle

\begin{abstract}
We introduce \href{https://github.com/numbbo/coco}{COCO}, an open source platform for Comparing Continuous Optimizers in a black-box setting.
\href{https://github.com/numbbo/coco}{COCO} aims at automatizing the tedious and repetitive task of
benchmarking numerical optimization algorithms to the greatest possible
extent.
The platform and the underlying methodology allow to benchmark in the same framework deterministic and stochastic solvers for both single and multiobjective optimization.
We present the rationals behind the (decade-long) development of the platform
as a general proposition for guidelines towards better benchmarking.
We detail underlying fundamental concepts of \href{https://github.com/numbbo/coco}{COCO} such as the definition of
a problem as a function instance, the underlying idea of instances,
the use of target values, and runtime \new{defined by the number of function calls} as the central performance measure.
Finally, we  give a quick overview of the basic
code structure and the currently available test suites.
\end{abstract}


\section{Introduction}
\label{\detokenize{index:introduction}}
We consider the continuous black-box optimization or search problem to minimize
\begin{equation}\label{eq1}
\begin{split}f: X\subset\mathbb{R}^n \to \mathbb{R}^m \qquad n,m\ge1\end{split}
\enspace,
\end{equation}
\new{where the search domain $X$ is typically a bounded hypercube or the entire continuous space.}\footnote{%
\new{We later also consider integer variables. They are embedded in the continuous space and labeled for the solver as integers,
see also Section~\ref{sec:test-suites}.
}}
\del{
such that for the \(l\) constraints
\begin{equation}\label{eq2}
\begin{split}g: X\subset\mathbb{R}^n \to \mathbb{R}^l \qquad l\ge0\end{split}
\end{equation}
we have \(g_i(\x)\le0\) for all \(i=1\dots l\).
}%
More specifically, we aim to find, as quickly as possible, one or several solutions \(\x\) in the search space \(X\) with \emph{small} value(s) of \(f(\x)\in\mathbb{R}^m\).
\del{ that satisfy
(or approximate) all constraints \(g\).}

\newcommand{\lbz}{\linebreak[0]}
A continuous optimization algorithm,
denoted as \emph{solver}, addresses the
above problem.
In this paper we only consider zero-order black-box optimization \cite{nemirovski1995information, bubeck2014convex, nesterov2018lectures}: while
the search domain \(X \subset\mathbb{R}^n\)
and its boundaries are accessible,
no other prior knowledge about \(f\) is available to the solver.\footnote{%
In the multiobjective case also the upper values of interest in $f$-domain are provided, see also Appendix~\ref{app:mo}.}
That is, \(f\)\del{ and \(g\) are} \new{is} considered as a black-box, also known as an oracle,
and the \emph{only} way the solver can acquire information on $f$ is by
querying the value
\(f(\x)\) of a solution \(\x\in X\).
Zero-order black-box optimization is thus a derivative-free optimization setting.\nnew{\footnote{
\nnew{%
In \cite{audet2017derivative}, blackbox optimization (BBO) is defined as ``{\it the study of design and analysis of algorithms that assume the objective and/or constraint functions are given by blackboxes}'',
and a blackbox is defined as ``{\it \dots any process that when provided an input, returns an output, but the inner workings of the process are not analytically available.
The most common form of blackbox is computer simulation, but other forms exist, such as laboratory experiments for example}''.
We prefer to define the black-box setup solely from the \emph{interfacing} between problem and solver and the exchange of information, which is
to a large degree independent of the underlying problem (e.g.\ its analytical nature or the availability of gradients).
In our case, the inner workings of the black-boxes \emph{are} analytically available to us,
but the solver is not allowed to access or use them.
}}}
\del{ This formalization of zero-order black-box optimization is in line with \cite{nesterov2018lectures,nemirovski1995information} where the functions $f$\del{ and $g$ are} \new{is} represented as oracles (see also \cite{bubeck2014convex}).}
We generally consider ``time'' to be the number of calls to the function \(f\) \new{and will define ``runtime'' correspondingly}.

\del{When \(l = 0\) \new{in Equation \eqref{eq2}}, we are in the common setting of unconstrained optimization.}
When \(m > 1\) \new{in Equation \eqref{eq1}}, we are in the setting of multiobjective optimization.
\new{Here, we only consider the case where $m\in\{1, 2\}$,
whereas the presented framework is in general designed to be extendable to other settings
(see also Sections~\ref{sec:test-suites} and~\ref{sec:extensions}).}
%\footnote{\new{
%For example, a test suite with various number of constraints is currently in development.
%}} % end footnote

From these prerequisites, benchmarking solvers seems to be a
rather simple and straightforward task. We run a solver on a collection of
problems and display the results. However, under closer inspection,
benchmarking turns out to be surprisingly tedious.
A set of objective functions has to be selected, problem instances should be derived,
an experimental design has to be established,
a set of performance measures has to be chosen,
data have to be recorded,
and results have to be exposed and interpreted in a comprehensive and comprehensible way.
Each of these steps asks for a great number of subtle decisions and is yet crucial for the
validity of the outcome (the chain is only as strong as its weakest link).
In particular, we require here to
get results that can be meaningfully interpreted beyond the
standard conclusion that on some problem some solver is better than another.%
\footnote{A common major flaw, unless data or performance profiles are used, is
to have no indication of \emph{how much} better a\ddel{ better} solver is.
That is, benchmarking results often provide no indication of
\emph{relevance}, for example, when
the main output consists of hundreds of tabulated numbers only interpretable on
an ordinal (ranking) scale.
This problem is connected to the common shortcoming of not clearly distinguishing
\emph{statistical significance} and \emph{relevance}.
Statistical significance is only a secondary and by no means sufficient condition for relevance.
} % end of footnote

The difficulty of proper experimental analysis and benchmarking of
solvers has already been recognized in previous work, see in
particular \cite{BAR1995,JOH2002,BHL2017}, which also give guidelines for better
experimental work.
In our work, we offer a conceptual guideline for benchmarking
continuous optimization algorithms addressing these challenges
that is implemented within the \href{https://github.com/numbbo/coco}{COCO} framework.%
\footnote{%
\href{https://github.com/numbbo/coco}{COCO} has been continuously developed since 2008. For implementation details,
confer to \href{https://www.github.com/numbbo/coco}{the code basis} on
\nnew{GitHub} and the \href{http://numbbo.github.io/coco-doc/C/}{C API documentation}.
} % end of footnote

\begin{figure}
  \centering
\noindent{\hspace*{\fill}\includegraphics[width=1.000\linewidth]{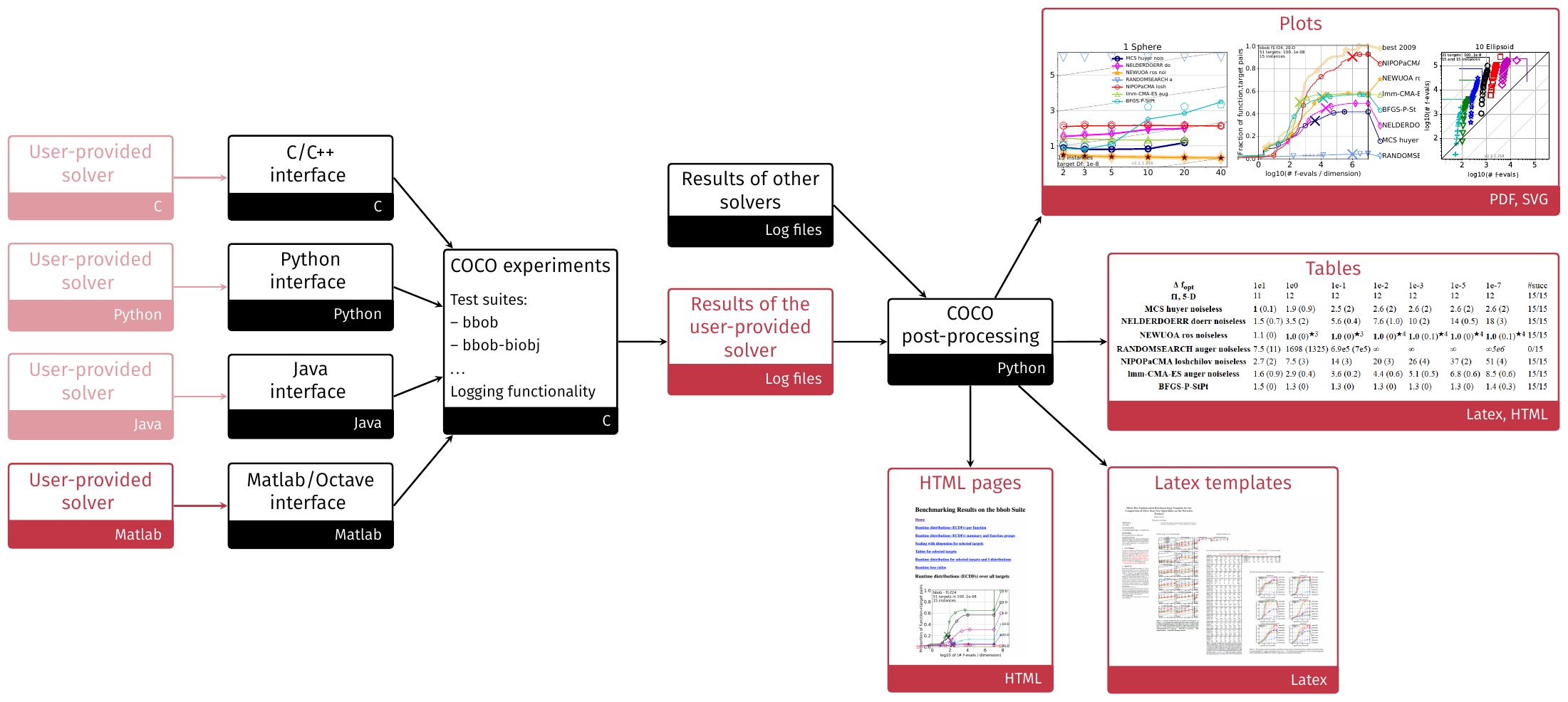}\hspace*{\fill}}
\caption{\label{fig:coco}
Overview of the \href{https://github.com/numbbo/coco}{COCO} platform.
\href{https://github.com/numbbo/coco}{COCO} provides all black parts while
users only have to connect their solver to the \href{https://github.com/numbbo/coco}{COCO} interface
in the language of interest, here for instance Matlab, and to decide on the
test suite the solver should run on. The other \new{red} components show the output
of the experiments \new{(number of function evaluations to reach certain target precisions)}
and their post-processing and
are automatically generated. For the \emph{results of other solvers}
part, \href{https://github.com/numbbo/coco}{COCO}
provides data from more than 200
pre\new{viously}-run benchmark experiments
with a large variety of solvers \new{collected over the last ten years from dozens of researchers}.
}
\end{figure}

The \href{https://github.com/numbbo/coco}{COCO} framework provides the following practical means for an automatized \new{black-box optimization}
benchmarking procedure (see also Figure~\ref{fig:coco}):
\begin{itemize}
\item {} 
an interface to several languages in which the benchmarked solver
can be written, currently C/C++, Java, Matlab/Octave and Python,

\item {} 
several suites of test problems, currently all written in C,
where each problem can assume an arbitrary number of pseudo-randomized instances,

\item {} 
data logging facilities,

\item {} 
data post-processing written in Python that produces various plots and tables,

\item {} 
\new{empirical} results of other solvers that can be used for comparison,\footnote{%
\new{\COCO\ provides a software environment for black-box optimization benchmarking but neither a server to run experiments nor the solvers themselves.}}

\item {} 
HTML pages assembling these plots and tables to ease their inspection,

\item {} 
LaTeX templates that include some selected results.

\end{itemize}
The underlying philosophy is to provide everything that experimenters
need to set up and implement when they want to benchmark a given solver
implementation \emph{properly}.
A desired side effect of reusing the same framework is that data collected
over years or even decades can be effortlessly compared.\footnote{%
See \href{http://coco.gforge.inria.fr/data-archive}{http://coco.gforge.inria.fr/data-archive}
\new{and {\ttfamily cocopp.archives} in the {\ttfamily cocopp} Python module}
to access all \new{collected} data, \new{in particular those} submitted
to the Black-Box Optimization Benchmarking (BBOB) \href{http://numbbo.github.io/workshops/}{\mbox{workshop series}} at the GECCO conference.
} % end of footnote
So far, the framework has been successfully used to benchmark over
200 different solvers by dozens of researchers. The data from all these experiments
are openly accessible and can be seamlessly used directly within the \href{https://github.com/numbbo/coco}{COCO}
post-processing as will be showcased later. These data come from solvers of different nature:
deterministic (pattern-search-based, trust-region derivative-free optimization, quasi-Newton, ...) and stochastic
(evolution strategies, Bayesian optimization, differential evolution, ...) addressing single- and multiobjective problems.

The purpose of this paper is to present the \href{https://github.com/numbbo/coco}{COCO} platform and give an overview of the main ideas it is based upon.
\new{%
The remainder of this section discusses related work 
and terminology that will be used later on. Section~2 discusses the motivations and objectives behind \COCO. Section~3 presents the central theses describing our approach to benchmarking methodology. Section~4 gives a summary of already available test suites. Section~5 presents practical examples how to use the \COCO\ software. Sections~6 and 7 show some usage statistics and extensions under development. Section~8 concludes the paper with a summary and discussion.
\new{The appendices provide details on how we chose test functions, the difference to competitive testing, and how we measure performance on biobjective functions.}

}

\subsection{Related Work}
\label{\detokenize{index:related-work}}
Benchmarking solvers is an important task in optimization,
independent of which type of problem is tackled. Experimental studies
reach back as far as the 1950s \cite{HMSW1953} and we refer to \cite{BHL2017} for a recent
summary of the history of benchmarking.
A few benchmarking  \emph{platforms} have been proposed over the years alongside \href{https://github.com/numbbo/coco}{COCO}
that also facilitate the task of analysing data from benchmarking experiments.

\href{http://www.gamsworld.org/performance/paver2}{Paver}, the Performance Analysis and Visualization for Efficient Reproducibility
software \cite{BUS2014} is offered in the context of the
\href{http://www.gamsworld.org/}{GAMS World initiative} as a collection of Python 
scripts and available on \href{https://github.com/coin-or/Paver}{\nnew{GitHub}}.
\href{http://www.gamsworld.org/performance/paver2}{Paver} allows to read in CSV-formatted files of numerical benchmarking experiments from an
arbitrary number of solvers and offers ``counting solver runs with certain
properties, computing mean values and quantiles of solver run attributes, and
performance profiles'' \cite{BUS2014}, in particular in an HTML-based format.

A similar analysis is supplied by the \href{https://optimizationbenchmarking.github.io} platform.
Developed as a generic visual tool for benchmarking on arbitrary
(combinatorial, numerical, \nnew{etc.}) problems, it provides data profiles and median
convergence plots for a set of benchmarking experiments from an arbitrary number
of solvers.%
\footnote{
The \href{https://optimizationbenchmarking.github.io}{optimizationBenchmarking.org} framework
allows to read in data from the \href{https://github.com/numbbo/coco}{COCO} platform and one of the
introductory examples displays the \href{https://github.com/numbbo/coco}{COCO} data from the BBOB-2013 workshop.
} % end of footnote
The software is written in Java and is available in a dockerized
version. In contrast to \href{http://www.gamsworld.org/performance/paver2}{Paver}, where the user invokes the analysis from the
command line, the user interacts with the \href{https://optimizationbenchmarking.github.io}{optimizationBenchmarking.org} framework
via the browser.

The latest software for analysing solver data is the
\href{https://github.com/IOHprofiler}{IOHProfiler} \cite{DWYV2018} which is inspired by the \href{https://github.com/numbbo/coco}{COCO} platform. Its post-processing part, in particular,
allows to visualize solver performance (given in \href{https://github.com/numbbo/coco}{COCO} format) interactively
in the browser and offers both fixed-target and fixed-budget views.

In the context of benchmarking multiobjective solvers, a few additional software
packages should be mentioned. \href{https://sop.tik.ee.ethz.ch/pisa}{PISA} presented
one of the first attempts to integrate the collection of data from benchmarking
experiments and their analysis. It
provides both multiobjective test problems (discrete and continuous; toy
as well as real-world problems) and an extensive set of scripts
to assess the solvers' performance in terms of quality indicator measures such as
the hypervolume indicator and various epsilon indicators \cite{BLTZ2003}. A \new{special feature}\del{specificity}
of PISA is the split of the solvers into a selector and a
variator part where the latter contains the variation operators for a
specific (set of) optimization problems.
\new{A disadvantage of PISA is that it has not been actively maintained for a long time.}

Newer\del{ and thus still supported} platforms that attempt to
benchmark multiobjective
solvers are the \href{http://moeaframework.org}{MOEA Framework} (in Java),
its Python counterpart \href{https://github.com/Project-Platypus/Platypus}{Platypus},
\href{https://github.com/jMetal/}{jMetal} (originally in Java, now also in other
languages) \cite{DN2011}, PLATEMO (in Matlab) \cite{TCZJ2017},
and ecr (in R) \cite{BOS2018}.
All these platforms focus on providing a large amount of solvers and a comprehensive
set of test functions \new{and are open source}. Performance assessment is typically restricted to
statistical tests and tabular visualizations of achieved
quality indicator values at certain budgets although certain
proprietary products such as \href{https://www.esteco.com/modefrontier}{ModeFRONTIER}
provide additional visual output.

In all these platforms, except for the \new{commercial ModeFRONTIER},
the entire benchmarking experiment is only semi-automated in
that the user still has to decide on the concrete benchmarking experiments
and all their intricacies. At most the data collection and visualization are automated.
\href{https://github.com/numbbo/coco}{COCO}, on the contrary, provides concrete benchmarking suites
and a predefined setup for the experiments in order to facilitate the
comparison of a large number of solvers. Providing the corresponding solver data sets
from \href{https://github.com/numbbo/coco}{COCO} experiments online%
\footnote{
See \url{http://coco.gforge.inria.fr/doku.php?id=algorithms}
} % end of footnote
and also directly through the \href{https://github.com/numbbo/coco}{COCO}
post-processing interface (see Section~\ref{sec:analyzing-the-results}) is another
unique aspect that sets \href{https://github.com/numbbo/coco}{COCO} apart.

Besides entire benchmarking platforms, many collections of test functions
have been proposed. Examples are the problems collected by
Moré, Garbow, and Hillstrom \cite{MGH1981}, Hock and Schittkowski \cite{HS1981},
Schittkowski \cite{SCHI1987}, \href{http://plato.la.asu.edu/bench.html}{Mittelmann},
\href{https://www.mat.univie.ac.at/~neum/glopt/test.html}{Neumaier}, the
\href{http://www.cuter.rl.ac.uk/Problems/mastsif.shtml}{CUTEr/st suite} or
the randomly generated, but structured problems from the GKLS generator \cite{GKLS2003}.
The \href{http://www.gamsworld.org/performance/performlib.htm}{GAMSWorld webpage}
lists further common problem suites.
Other sets of test functions to mention are those used for the
\href{http://www.ntu.edu.sg/home/EPNSugan/index\_files/cec-benchmarking.htm}{competitions at the CEC conferences}
since 2005.
They evolved quite a bit over the years and nowadays use the same technique of
problem transformations as introduced with the \bbob functions of
the \href{https://github.com/numbbo/coco}{COCO} platform.

Several test suites with multiobjective problems have also been proposed, of which several
possess questionable properties, see
\cite{BRO2019} for a discussion.

The majority of the mentioned test suites contain a large proportion of
simple-to-solve and non-scalable problems without instance variations.
Aggregating the performance
over all functions of such a test suite and interpreting the results must
therefore be done with care---because the distribution of
function difficulties in a benchmarking experiment clearly determines which
solvers will excel.
Alternatively, \href{https://github.com/numbbo/coco}{COCO} aims to implement suites with balanced difficulties observed in practice and with a bias towards difficult-to-solve functions.

\subsection{Terminology}
\label{\detokenize{index:terminology}}
We specify a few terms which are used later.
\begin{description}
\item[{\emph{function}}] \leavevmode
We talk about an objective \emph{Function} as a parametrized mapping
\(\mathbb{R}^n\to\mathbb{R}^m\) with scalable input space, \(n\ge1\),
and usually \(m\in\{1,2\}\) (single- and bi-objective).
Functions are parametrized such that different \emph{instances} of the
``same'' function are available, e.g., translated or shifted versions.

\item[{\emph{quality indicator measure}}] \leavevmode
From the \new{function} \del{\(f\) and \(g\)-}values of a \new{set of} solutions we compute a scalar \emph{Quality indicator}
value which is used in the performance assessment.
By convention and abuse of naming, the ``quality'' indicator is minimized.
The indicator assigns to any solution \emph{set} a real value.
\new{Our} simplest quality indicator measure is the minimum of all so-far-observed \(f\)-values.
In the multiobjective case, a typical example is the hypervolume of all
so-far-observed solutions
(where the position of the reference point is a further design decision to be
made).\del{
In the constrained case we suggest the minimum of all so-far-observed
\(f + 1000\sum_{i=1}^l g_i 1\!\!1\{g_i>0\}\)-values.}

\item[{\emph{problem}}] \leavevmode
We talk about a \emph{Problem}, as a specific \emph{function
instance} on which a solver is run.
A problem can be evaluated for a solution \(\x\in\mathbb{R}^n\) and returns\del{ the values}
\(f(\x)\)\del{ and \(g(\x)\)}.
In the context of performance assessment, a target value
is added to define a problem. A problem is considered as
solved when the quality indicator value reaches this target,
which is \new{always}\del{usually} given as a precision value, i.e., as a deviation from a
(supposedly) optimal value.

\item[{\emph{runtime}}] \leavevmode
We define \emph{Runtime} or \emph{run-length} \cite{HOO1998} as the \emph{number of
function evaluations} or \(f\)-evaluations conducted on a given problem until a prescribed target value is hit.\del{
In the constrained case we typically count \(f\) and \(g\)-evaluations equally.}
Runtime is our central performance measure.

\item[{\emph{suite}}] \leavevmode
A test or benchmark \emph{Suite} is a collection of problems in different dimensions.
It typically consists of 1000 to 5000 problems
(number of dimensions $\times$ number of functions $\times$ number of instances),
where the number of objectives \(m\) is fixed.

\end{description}

\section{Why COCO?}
\label{sec:why-coco}
Our aim in providing the \href{https://github.com/numbbo/coco}{COCO} platform is \new{three}\del{two}fold:
\begin{itemize}
\item\del{ {} to}
diminish the time burden, the pitfalls, the bugs
and the omissions of the repetitive coding task of setting up and running a benchmarking
experiment,

\item\del{ {} to}
provide a \emph{conceptual guideline for better benchmarking}, \new{and}

\item \new{provide a growing archive of comparative benchmarking data to the scientific community.}

\end{itemize}

Our setup \new{has a distinct boundary between
the implementation of benchmark functions on the one hand
and the experimental design, data collection and presentation
on the other hand. Our benchmarking}
guideline has the following defining features.
\begin{enumerate}
\item {} 
We benchmark solvers on a set (a suite) of
benchmark functions.
Benchmark functions are
\begin{enumerate}
\item {} 
used as black boxes for the solver, however, they
are explicitly known to the scientific community,

\item {} 
designed to be comprehensible, to allow a meaningful
interpretation of performance results,

\item {} 
difficult to ``defeat'' or exploit, that is, they \new{should}\del{do} not
have artificial regularities and \new{artificial} symmetries that can easily be exploited (intentionally or unintentionally),%
\footnote{
For example, the global optimum is not in all-zeros, optima are
not placed on a regular grid, most functions are not separable \cite{WHI1996}.
The objective to remain comprehensible makes it more challenging to design
non-regular functions. Which regularities are common place in real-world
optimization problems remains an open question.
} % end of footnote

\item {} 
scalable with the input dimension \cite{WHI1996},

\item {} 
instantiated from an arbitrary number of pseudo-randomized versions, i.e, instances,

\item {} 
models for ``real-world'' problems; the currently available test suites
(see Section~\ref{sec:test-suites}) model in particular well-known problem \emph{difficulties} like
ill-conditioning, multimodality and ruggedness.

\end{enumerate}

The input parameters to the solver must not depend on the specific function within the benchmark suite. They can, however, depend on the black-box \emph{interface} (i.e., the signature of the function), namely on dimension\del{, number of constraints,} and search domain of interest (e.g., to set variable bounds).

\item {} 
There is no predefined budget (number of \(f\)-evaluations) for running an
experiment, the experimental procedure is \emph{budget-free} \cite{HAN2016ex}.
Specifically, all results are comparable irrespectively of the chosen budget
and up to the smallest budget in the compared results. Hence, the larger
the budget, the more data are generated to compare against. The smaller
the budget, the less meaningful conclusions are possible (which becomes
most evident when the budget approaches zero).
This also implies an anytime assessment approach\new{: 
the performance is not (only) measured after some given
runtime or fixed budget or after reaching some given target
but over the entire run of the solver}.

\item {} 
\emph{Runtime}, measured in number of \(f\)-\del{ and/or \(g\)-}evaluations \cite{HAN2016perf},
is the only used performance measure.
It is further aggregated and displayed in several ways. The
advantages of runtime are that it
\begin{itemize}
\item {} 
is independent of the computational platform, language, compiler, coding
style\del{s}, and other specific experimental conditions%
\footnote{
Runtimes measured in \(f\)-evaluations are widely
comparable and designed to stay. The experimental procedure
includes, however, also a timing experiment which records the
internal computational effort of the solver in CPU or wall clock time \cite{HAN2016ex}.
},

\item {} 
is independent, as a measurement, of the specific function on which it has
been obtained, \new{that is, \emph{``taking 42 evaluations''} has the same meaning on any function,
while \emph{``reaching a function value of 42''} has not,}

\item {} 
is relevant, meaningful and easily interpretable without expert domain knowledge\new{,}

\item {} 
is quantitative on the ratio scale%
\footnote{
As opposed to a ranking of solvers based on their solution quality
achieved after a given budget of evaluations.
} % end of footnote
\cite{STE1946},

\item {} 
assumes a wide range of values, and

\item {} 
aggregates over a collection of values in a meaningful way%
\del{\footnote{
With the caveat that the \emph{arithmetic average} is dominated by large values
which can compromise its informative value.
}}.

\end{itemize}

A \emph{missing} runtime value is considered as a possible outcome (see Section~\ref{sec:runtime-and-target-values}).

\item {} 
The display of results is done with the distinct effort to be as comprehensible,
intuitive and informative as possible.
In our estimation, details can matter a lot.

\end{enumerate}

We believe, however, that in the \emph{process} of solver \emph{design}, a benchmarking
framework like \href{https://github.com/numbbo/coco}{COCO} has its limitations.
During the design phase, usually
\begin{itemize}
\item {} 
fewer benchmark functions should be used,

\item {} 
the functions and measuring tools should be tailored to the given solver
and the design question, and

\item {} 
the overall procedure should be more informal and interactive with rapid iterations.

\end{itemize}

A benchmarking framework then serves to conduct the formalized validation
experiment of the design \emph{outcome} and can be used for regression testing.
Johnson \cite{JOH2002} and Hooker \cite{HOO1995} provide
excellent discussions of how to do experimentation beyond the specific
benchmarking scenario.

\section{Benchmarking methodology}
\label{\detokenize{index:benchmarking-methodology}}
This section details the benchmarking methodology used in \href{https://github.com/numbbo/coco}{COCO}
which, in particular,
allows for a budget-free experimental design.
We elaborate on functions, instances, problems, runtime, target values, restarts, simulated restarts, performance aggregation, and a budget-dependent benchmarking setup.

\subsection{Functions, Instances, and Problems}
\label{\detokenize{index:functions-instances-and-problems}}
In the \href{https://github.com/numbbo/coco}{COCO} framework we consider \textbf{functions}, \(f_i\), for each suite
distinguished by their identifier \(i=1,2,\dots\) .
Functions are further \emph{parametrized} by the (input) dimension, \(n\), and the
instance number, \(j\).
We can think of \(j\) as an index to a continuous parameter vector setting.
It parametrizes, among other things, search space translations and rotations.
In practice, the integer \(j\) identifies a single instantiation of these parameters.
For a given \(m\), we then have
\begin{equation*}
\begin{split}\finstance_i \equiv f\new{[}n, i, j\new{]}:\R^n \to \mathbb{R}^m \quad
\x \mapsto \finstance_i (\x) = f\new{[}n, i, j\new{]}(\x)\enspace.\end{split}
\end{equation*}
Varying \(n\) or \(j\) leads to a variation of the same function
\(i\) of a given suite.
Fixing \(n\) and \(j\) of function \(f_i\) defines an optimization \textbf{problem}
\((n, i, j)\equiv(f_i, n, j)\) that can be presented to the solver.
Each problem receives again an index within the suite,
mapping the triple \((n, i, j)\) to a single number.

As this formalization suggests, the differentiation between function
and instance index is of purely semantic nature.
This semantics is, however, important in how we interpret and display results.
We interpret \textbf{varying the instance} parameter as
a natural randomization for experiments%
\footnote{
Changing or sweeping through a relevant feature of the problem class,
systematically or randomized, is another possible usage of instance
parametrization.
} % end of footnote
in order to
\begin{itemize}
\item {} 
generate repetitions on a single function for deterministic solvers,
making deterministic and non-deterministic solvers \new{directly}
comparable \new{(both are benchmarked with the same experimental setup)\footnote{%
\new{The number of instances and the number of runs per instance may be varied.
By default, all solvers execute one run per instance and instances are interpreted as repetitions. We initially used a setup with 3 runs per instance for stochastic solvers which would allow to distinguish within- and between-instance variance. Either setup is available to the user and compatible with the data processing and aggregation.
}}},

\item {} 
average away irrelevant aspects of the function definition,

\item {} 
alleviate the problem of overfitting, and

\item {} 
prevent exploitation of artificial function properties,

\end{itemize}
thereby providing a ``fairer'' and more robust setup.
For example, we do not consider the absolute position of the optimum
as a defining function feature.
Therefore, in a typical \href{https://github.com/numbbo/coco}{COCO} benchmark suite, instances with different (pseudo-randomized) search space translations are presented to the solver.
If a solver is translation invariant (and hence ignores domain
boundaries), this is equivalent to varying the initial solution.

\subsection{Runtime and Target Values}
\label{sec:runtime-and-target-values}
In order to measure the runtime of a solver on a problem, we
consider a hitting time condition.
We define a non-increasing \emph{quality indicator measure} and prescribe a set of \textbf{target values}, \(t\) \cite{HAN2016perf,BRO2015,BRO2016}.
In the single-objective unconstrained case, the quality indicator is the best so-far-seen \(f\)-value.%
\footnote{
In the constrained, multiobjective and noisy cases, the quality indicator measure is more intricate.
In the multiobjective case, for example, we use a version of the hypervolume indicator
of all so-far evaluated solutions
and approximate the optimal value from experimental data in order to derive
informative target values \cite{BRO2016}. For more information, please see Appendix~\ref{app:mo}.
} % end of footnote
For a single run, when the quality indicator reaches or surpasses a target value \(t\)
on problem \((f_i, n, j)\), we say the solver has \emph{solved the problem} \((f_i, n, j, t)\)---the solver was successful.%
\footnote{
Reflecting the \emph{anytime} aspect of the experimental setup,
we use the term \emph{problem} in two meanings: (a) for the problem the
solver is benchmarked on, \((f_i, n, j)\), and (b) for the problem, \((f_i, n, j, t)\), a solver may
solve by hitting the target \(t\) with the runtime, \(\mathrm{RT}(f_i, n, j, t)\), or may fail to solve \new{within the experimentation budget}.
Each problem \((f_i, n, j)\) gives raise to a collection of \emph{dependent} problems \((f_i, n, j, t)\).
Viewed as random variables, given \((f_i, n, j)\), the \del{events} \(\mathrm{RT}(f_i, n, j, t)\) are not
independent \del{events} for different values of \(t\).
In particular, the Cumulative Distribution Function (CDF) for any larger \(t\) dominates the CDF for any smaller
(i.e., more difficult) \(t\).
} % end of footnote
 \new{We typically collect runtimes for around a hundred different target values from each \emph{single} run.}

Target values are directly linked to a problem, and we leave the burden of
defining the targets with the designer of the benchmark suite.%
\footnote{
The alternative, namely to present the obtained \(f\)- or indicator-values as results,
leaves the (often rather insurmountable) burden to interpret the meaning of the
indicator values to the experimenter or the final audience.
Fortunately, budget-based targets are a generic way to generate target values
from observed runtimes as\del{ will be} discussed \new{at the end of the section}\del{shortly}.
} % end of footnote
More specifically, we consider the problems \((f_i, n, j, t(\new{i}, j))\) for
all \new{functions \(i\) and} benchmarked instances \(j\). The targets \(t(\new{i}, j)\) depend on the \new{problem} instance
in a way to make\del{ the} \new{certain} problems comparable.
\new{%
We typically define\del{d} the (absolute) target values from a single set of \emph{precision} values added to a reference offset which is the known (or estimated) optimal indicator value.
}\new{%
Runtimes from the same subset of target precision values are aggregated over problems or functions.
Only the precision values and not the \new{(by themselves meaningless)} target values are exposed to the user.
}

The \textbf{runtime} is the evaluation count when the target value \(t\) was
reached or surpassed for the first time.
That is, runtime is the number of \(f\)-evaluations\del{ and/or \(g\)-evaluations} needed to solve the problem
\((f_i, n, j, t)\).
\emph{Measured or numerically bootstrapped runtime values are in essence the only way
how we assess the performance of solvers.}

If a solver does not hit the target \(t\) in a given single run,
the run is considered to be \emph{unsuccessful}.
The runtime of this single run remains undefined,
but is bounded from below by the number of evaluations conducted during the run.
Naturally, increasing the budget that the solver is allowed to use
increases the number of successful runs and hence the number of available runtime measurements.
Therefore, larger budgets are preferable. However, larger budgets should not come at
the expense of abandoning prudent termination conditions.
Instead, restarts should be conducted (see also Section~\ref{sec:restarts-and-simulated-restarts}).

As an alternative to predefined target precision values, we also propose
to select target values based on the runtimes of a set of solvers \new{similar to data profiles}.
For any \emph{given budget}, we select the associated target in the following way \cite{HAN2016perf}:
from the finite set of recorded target values, we pick the easiest (i.e., largest) target for which
the expected runtime of all solvers (ERT, see Section~\ref{sec:aggregation}) exceeds the budget.%
\footnote{
\new{That is, we take the best solver as reference to compute budget-based targets.} Instead of \new{the best}\del{all} solver,
we could also take the median or (least promising) the worst solver or pure random search as reference.
} % end of footnote
The resulting target value depends on a set of solvers and on the given function and dimension.
Starting with a set of budgets of interest (e.g., relevant budgets for an application of interest),
we compute \new{this way}\del{like this} a set of \textbf{budget-based target values} (AKA run-length-based targets) for each function and dimension.
This choice of target values does not require any knowledge about the underlying functions and their indicator values, but it requires experimental data and depends on the chosen set of solvers.

\subsection{Restarts and Simulated Restarts}
\label{sec:restarts-and-simulated-restarts}
Any solver is bound to terminate and, in the single-objective case,
return a single recommended solution, \(\x\), for the problem, \((f_i, n, j)\).
More generally, we assume an anytime and any-target scenario,
considering a non-increasing quality indicator value computed in each time step
from all evaluated solutions.
Then, at any given time step, the solver solves all problems \((f_i, n, j, t)\) for which the current indicator value hit or surpassed the target \(t\).

Independent restarts from different, randomized initial solutions are a simple
but powerful tool to increase the number of solved problems \cite{HAR1999}---namely by increasing the number of \(t\)-values, for which the problem \((f_i, n, j)\) was solved.
Increasing the budget by independent restarts will increase the success rate, but generally does not change the performance assessment in our methodology,
because the additional successes materialize at higher runtimes \cite{HAN2016perf}.
Therefore, we call our approach \emph{budget-free}.
Restarts, however, ``\emph{improve the reliability, comparability, precision, and `visibility' of the measured results}'' \cite{HAN2016ex}.

\textbf{Simulated restarts} \cite{HAN2010,HAN2016perf} are used to determine a runtime for unsuccessful runs.
Semantically, simulated restarts are only meaningful if we can
interpret different instances as \emph{random repetitions} that could arise,
for example, by restarting from different initial solutions on the same instance (hence we do not use simulated restarts for the multiobjective case).
Resembling the bootstrapping method \cite{EFR1994} when we face an unsuccessful run,
we draw further runs from the set of tried problem instances,
uniformly at random with replacement, until we find an instance, \new{$j$}, for which \((f_i, n, j, t\new{(i, j)})\) was solved.
The evaluations done on the first\del{ unsolved problem} and on all subsequently
drawn unsolved problems are added to the runtime on the last, solved problem and
are considered as the runtime on the originally unsolved problem instance.
This method is only applicable if at least one problem instance was solved and is applied if at least one problem instance was not solved.
By their nature, the success probability of ``runs'' with simulated restarts is
either zero (if no problem instance was solved) or one.
\emph{Simulated restarts allow to directly compare solvers with vastly different
success probabilities.}

\subsection{Aggregation}
\label{sec:aggregation}
A typical benchmark suite consists of about 20--100 functions with 5--15 instances for each function. For each instance, up to about 100 targets are considered for the
performance assessment. This means we consider at least \(20\times5=100\), and
up to \(100\times15\times100=150\,000\) runtimes to assess performance.
To make them amenable to examination and interpretation, we need to summarize these data.

The semantic idea behind aggregation is to compute a statistical summary over a
set or subset of problems of interest \emph{over which we assume a uniform
distribution}.
From a practical perspective, this means we assume to face each problem with similar probability
\emph{and} we have no simple way to distinguish between these problems
to select a solver accordingly.
If we can distinguish between problems easily, for example, according to their input dimension, the aggregation of data (for a single solver) is rather counterproductive.
Because the dimension is known in advance and can be used for solver selection, we never aggregate over dimension.
This has no significant drawback when all functions are scalable in the dimension.

We use several ways to aggregate the measured runtimes.
\begin{itemize}
\item {} 
Empirical cumulative distribution functions of runtimes (runtime ECDFs), 
also denoted as (empirical) runtime distributions.
In the domain of
numerical optimization, ECDFs of runtimes to reach a given
\emph{single} target \new{precision} value
are well-known as \emph{data profiles} \cite{MOR2009}. Performance profiles are ECDFs
of these runtimes relative to the respective best solver \cite{DM2002}. We
favour absolute runtime distributions over performance profiles for two reasons:
\begin{enumerate}
\item {} 
Runtime ECDFs are unconditionally comparable across different publications \cite{JOH2002}.\new{\footnote{%
While runtime distributions by themselves are comparable across different publications,
this is generally not the case for data profiles.
The latter compute the target value for each problem individually
as $f_\text{best} + \tau (f_0 - f_\text{best})$,
based on the best achieved $f$-value from a set of solvers
within a given budget
and for the precision parameter $\tau$ \cite{MOR2009}.
}} %footnote
They are absolute performance measures (opposed to relative measures) and\new{, in our case,} do not
depend on other solvers for normalization.%
\footnote{
This advantage\del{ over performance profiles} partly disappears with
budget-based target values, as considered in Section~\ref{sec:runtime-and-target-values}.
Conceptually, performance profiles change the
displayed measurement potentially depending on all displayed solvers.
Budget-based target values,
\new{similar to the approach in data profiles,}
change the considered problems,
namely the target precisions which define when a problem is solved,
based on the performance of a set of solvers.
Budget-based target values
make the \new{unconditionally comparable but} somewhat arbitrary targets-to-reach setting somewhat less arbitrary.
}
 Performance profiles suffer from the deficiency of any relative
comparison procedure:
adding or removing a single entry may significantly affect the result of
other pairwise comparisons \cite{MER2015,GOU2016}.

\item {} 
Runtime ECDFs allow to distinguish easy problems from difficult problems (for
any given solver).
In performance profiles,
a small runtime ratio gives no indication of the problem difficulty
at all and a large runtime ratio can still mainly be caused by
a competitor which solves the problem very quickly.
An easy-versus-difficult problem classification is, however, a powerful feature
to select the best applicable solver \cite{HAN2010}.

\end{enumerate}

We usually aggregate not only runtimes from a single target \new{precision} value,
as in data profiles, but from several targets per function.
Because we display the x-axis on a log scale, the area
above the curve and the \emph{difference area} between two curves are meaningful
notions even when results from several problems are shown in a single
graph.
The geometric average runtime difference
\newcommand{\nb}{{\new{k}}}
\begin{equation}
\begin{split}\exp\left( \frac{1}{\nb}\sum_{i\new{=1}}^\nb \log\left(\frac{B_i}{A_i}\right)\right) =
\exp\left( \frac{1}{\nb}\sum_{i\new{=1}}^\nb \log(B_i) - \frac{1}{\nb}\sum_{i\new{=1}}^\nb \log(A_i)\right)\end{split}\label{eq:geo}
\end{equation}
reflects this difference area between the graphs in log-display and is invariant under re-sorting \new{of data}.

\new{The runtime distribution on a single problem over all targets
is tightly related to the convergence graph that displays the so far best
$f$-value against the number of $f$-evaluations. Consider the convergence graph with reversed y-axis (like for maximizing~\,$-\hspace{-0.5pt}f$). Then the runtime distribution is a step function strictly below (but close to)
this reversed convergence graph,
where the maximal y-distance results from the target discretization.
Hence, runtime ECDFs provide a single formalization for convergence graph data (when many targets are used) and data profiles (when a single target \new{precision} is used) with a smooth transition between the two.
}

By sorting the runtime values \new{taken from several problems}, runtime ECDFs
disguise relevant information which can lead to a misinterpretation when one
graph dominates another. Domination in each point of a data profile does not
imply equal or better performance on \emph{each problem} that is presented in
the data profile. For example, consider the \new{two} runtimes on two problems\new{, respectively,}
to be 50 \new{and} 500 \new{for} \new{S}olver~A, and 1000 \new{and} 100 for \new{S}olver~B.
On \new{the first problem, Solver~A is 20 times faster than Solver~B. On} the second problem, \new{S}olver~B is 5 times faster than \new{S}olver~A. Yet,
in the \emph{sorted} data profiles, \new{Solver A} {(}50, 500{)} dominates \new{Solver B} {(}100, 1000{)} everywhere by a factor of two.

\item {} 
Averaging, as an estimator of the expected runtime.
The estimated expected runtime of the restarted solver, ERT,
is often plotted against dimension to indicate scaling with dimension.
The ERT, also known as Enes \cite{PRI1997} or SP2 \cite{AUG2005} or aRT, is computed by dividing
the sum of all evaluations before the target was reached
(from successful and unsuccessful runs) by the number of
runs that reached the target.
If all runs reached the target, this is the (plain) average number of evaluations
to reach the target.
Otherwise, the unsuccessful runs are fully taken into account as if restarts had
been conducted until the target was reached.
Like simulated restarts, ERT integrates out the observed success rate in the data.
This also answers the question of how to weigh failure rate and/or
solution quality versus speed \cite{BUS2014} with a semantically and
practically meaningful measure%
\footnote{
A practically less meaningful \new{but easier to obtain} ad hoc measure, known as Q-measure \cite{PRI1997}
or SP1 \cite{AUG2005},
uses only successful runs to compute an analogous ratio.
This can be useful to get a quick estimate of ERT without
the need to bother with effective termination conditions,
but is currently not used within \href{https://github.com/numbbo/coco}{COCO}.
}.
The \emph{arithmetic} average is only meaningful if the underlying distribution of
the values is similar and has light tails.
Ideally, the values stem from the same distribution.
Otherwise, the average of log-runtimes, \new{that is the}\del{or} \emph{geometric} average \del{(see the previous point),}\new{\eqref{eq:geo} or a shifted geometric mean \cite{georges2018feature}}
are feasible alternatives.

\item {} 
Simulated restarts, see Section~\ref{sec:restarts-and-simulated-restarts},
aggregate data from several runs into a ``single run'' to
supplement a missing runtime value, similarly as in the computation of ERT.
The same data can be used to simulate many runtimes via simulated restarts.
These runtimes are usually plotted as empirical cumulative distribution function.
The ERT is the expected runtime of these simulated restarts.

\end{itemize}

\subsection{Budget-Dependent Benchmarking}
\label{\detokenize{index:budget-dependent-benchmarking}}
The performance of some solvers depends on the total budget of function evaluations
given as an additional parameter to the solver.
Consider, for example, a hybrid solver that couples an explorative strategy with a local search method. A number of final function evaluations is typically reserved to additionally improve the best solutions \cite{MIH2006}.
Therefore, a single performance assessment of such solvers cannot be expected to faithfully predict
their performance for budgets that are different from the one that was used in the experiments.
The budgeted (non-anytime) performance of a budget-dependent solver can be better estimated by
repeatedly running the solver with increasing budgets \cite{TUS2017}.
This overestimates the performance (underestimates the runtime),
as if the ``optimal'' budget were known in advance
and given as parameter to the solver.
Depending on the number and size of the budgets, this can take a significant amount of extra time (in the worst case, the overhead is quadratic in the maximal budget).
By using budgets that are equidistant on the logarithmic scale, however, the time complexity of the overhead depends linearly on the maximal budget, making the approach usable in practice \cite{TUS2017}.

\section{Test Suites}
\label{sec:test-suites}
\new{An important feature of the \COCO\ framework is that new test suites can be added with comparatively little effort, thereby getting all the benefits of the established benchmarking setup for the new suite.}
Currently, the \href{https://github.com/numbbo/coco}{COCO} framework provides \new{the following}\del{six different} test suites \new{(listed in order of their introduction)}:
\begin{description}
\item[{\bbob}]
contains 24 functions in dimensions 2, 3, 5, 10, 20 and 40 in five subgroups: separable, moderate, ill-conditioned, multi-modal weakly structured, multi-modal with global structure \cite{HAN2009fun}.

\item[{\bbobnoisy}]
contains 30 noisy functions in dimensions 2, 3, 5, 10, 20 and 40 in three subgroups with three different noise models \cite{HAN2009noi}. \new{The code for this test suite is only available at\del{,
only implemented in the old code basis} \href{http://coco.gforge.inria.fr/doku.php?id=downloads}{coco.gforge.inria.fr}.}

\item[{\bbobbiobj}]
contains 55 bi-objective functions in dimensions 2, 3, 5, 10, 20 and 40 in 15 subgroups \cite{BRO2019}.

\new{%
\item[{\bbobbiobjext}]
contains 92 bi-objective functions in dimensions 2, 3, 5, 10, 20, and 40, including all 55 \bbobbiobj{} functions \cite{BRO2019}. \new{For the 37 new functions in the suite, the reference target values are not yet established.}
}

\item[{\bboblargescale}]
contains 24\del{ large-scale} functions in dimensions 20, 40, 80, 160, 320 and 640 and the same five subgroups as the \bbob suite \cite{VAR2018}.

\item[{\bbobmixint}]
contains 24 mixed-integer single-objective functions in dimensions 5, 10, 20, 40, 80 and 160 and the same five subgroups as the \bbob suite \cite{TUS2019}.
\new{%
Integer variables are embedded into the continuous space, that is, the problems can be evaluated as continuous problems and appear in the integer variables as piecewise constant functions.
}

\item[{\bbobbiobjmixint}]
contains 92 mixed-integer bi-objective functions in dimensions 5, 10, 20, 40, 80 and 160 and the same 15 subgroups as the \bbobbiobj suite \cite{TUS2019}.

\end{description}

Test suites are crucial in that they
ultimately define the anticipated purpose
of the benchmarked solvers.
For example, if we care about solving difficult problems more than about solving
easy problems, test suites must contain more difficult than easy problems, etc.

The listed test suites were designed with the remarks from Section~\ref{sec:why-coco} in mind.
In particular, they introduce a number of transformations in \(\x\)- and \(f\)-space
in order to make the functions less regular and less symmetrical and hence less susceptible to exploits of the relatively simple underlying formulas.

\new{The continuous variables of t}he functions are\del{ generally} unbounded.
The single-objective suites, however, guarantee that the global optimum is in a \new{bounded}\del{finite} domain that is known to the solver,
hence bounded solvers can be, and frequently have been, benchmarked as well.

\href{https://github.com/numbbo/coco}{COCO} allows to integrate new problem suites.
\new{The interface to integrate suites with arbitrary constraints is implemented and fully functional.}\footnote{\raggedright%
See \href{https://github.com/numbbo/coco/blob/master/howtos/create-a-suite-howto.md}{https://github.com/numbbo/coco/blob/master/howtos/create-a-suite-howto.md} and \href{https://github.com/numbbo/coco/blob/master/code-experiments/src/coco.h\#L306}{{https://github.com/numbbo/coco/blob/master/code-experiments/src/coco.h}} for more details.}
In order to play well with the performance measurement methodology,
new test suites should feature
\begin{itemize}
\item {} 
definitions of all functions in various dimensions,

\item {} 
pseudo-randomized function instances, for example, with different locations of the global optimum,

\item {} 
the same target precision levels for all function instances (as used by the logger, however, disguised for the solver).
A simple way to control the target levels function-wise is to multiply each ``raw'' function
by a different positive scalar.

\item \new{In case of noisy (stochastic) functions, the $f$-distribution of any two solutions should either be the same or one distribution should dominate the other. This way, any ambiguity as to which solution is better can be avoided.
}

\end{itemize}

In the performance evaluation, the defined target levels become part of the problem definition (see Section~\ref{sec:runtime-and-target-values}).

\section{Usage and Output Examples}
\label{sec:usage-and-output-examples}
The \href{https://github.com/numbbo/coco}{COCO} platform implementation consists of two major parts (see also
Figure~\ref{fig:coco}):
\begin{description}
\item[{The \emph{experiments} part}] \leavevmode
defines test suites, allows to conduct experiments, and provides the output data.
The \href{http://numbbo.github.io/coco-doc/C}{code base is written in C}.
\href{https://github.com/numbbo/coco}{COCO} amalgamates as a pre-compilation step all C source code files into the files {\ttfamily coco.h} and {\ttfamily coco.c}.
These two files suffice to compile and link to run all experiments.
Interface wrappers for further languages are provided,
currently for Java, Matlab/Octave, and Python.

\item[{The \emph{post-processing} part}] \leavevmode
processes the data and displays the benchmarking result.
This is the central part of \href{https://github.com/numbbo/coco}{COCO}.
The code is written in Python and heavily depends on \href{http://matplotlib.org/}{{\ttfamily matplotlib}}
\cite{HUN2007}, which might change in future versions. The post-processing
part of \href{https://github.com/numbbo/coco}{COCO} allows to compare experiments from multiple solvers,
in particular, from hundreds of data sets provided by the framework.

\end{description}

Examples of using the two parts are presented in the following two sections.

\subsection{Running an Experiment}
\begin{figure}\vspace{-3ex}  % 544
\centering
\noindent{\hspace*{\fill}\includegraphics[width=0.840\linewidth]{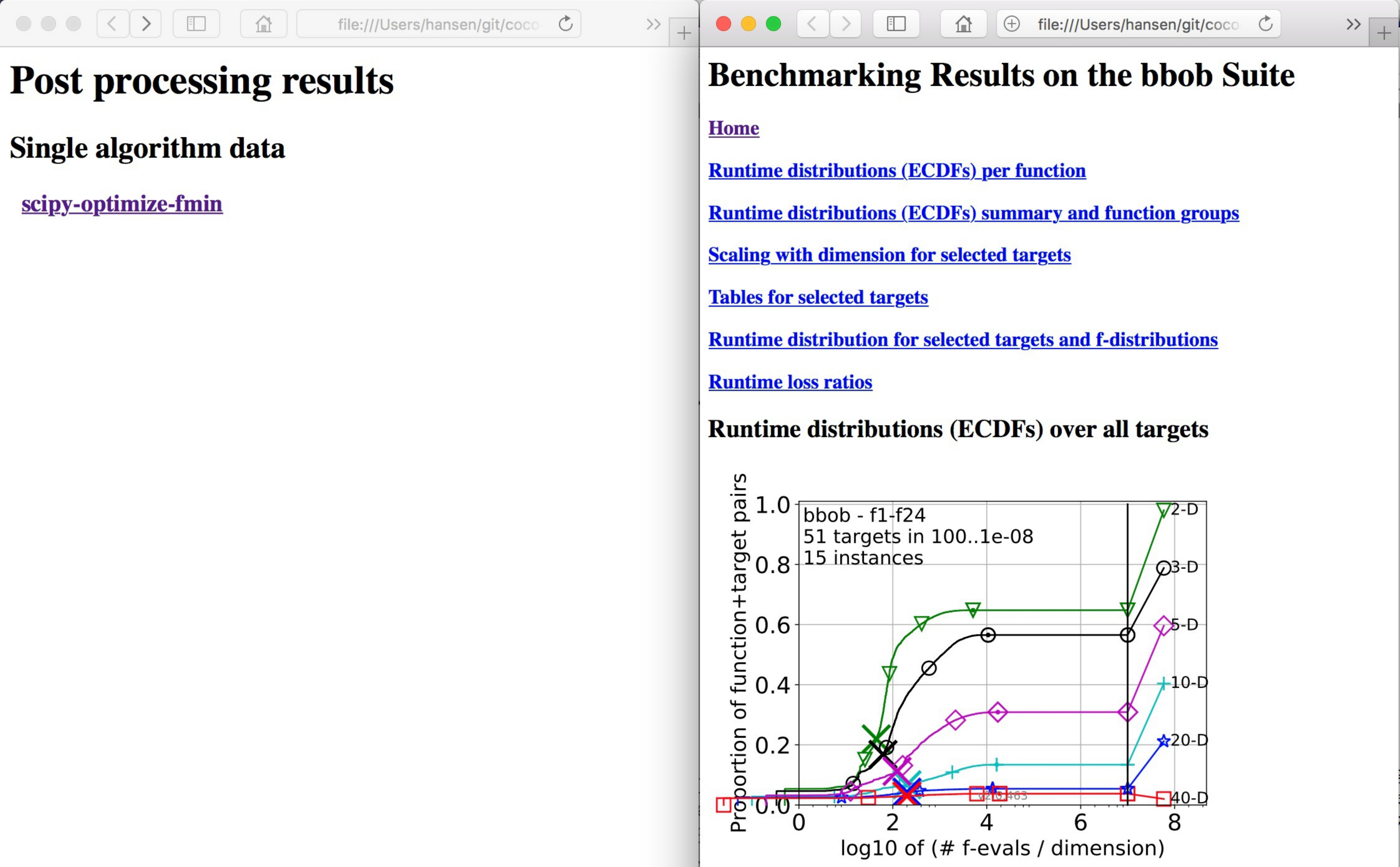}\hspace*{\fill}}
\caption{\label{fig:browseroutput}
Benchmarking output in the browser. Left: homepage \new{which opens when the post-processing has finished}; right: \new{page when clicking the \underline{\color{blue}{scipy-optimize-fmin}} link on the left. The further links on the right open various types of graphs most of which are discussed below.}\del{Single solver
data}
}
\end{figure}

\label{\detokenize{index:running-an-experiment}}
Installing \href{https://github.com/numbbo/coco}{COCO} in a linux system shell
(with the prerequisites of git and Python installed)
and benchmarking a
solver in Python, say, the solver {\ttfamily fmin} from {\ttfamily scipy.optimize},
is as simple as shown in Figure~\ref{fig:cococode},
\new{where the installation sequence is given above,
an example to invoke the benchmarking from a shell is given in the middle,
and a Python script for benchmarking is shown below}.%
\footnote{
See also \href{https://github.com/numbbo/coco/blob/master/code-experiments/build/python/example\_experiment\_for\_beginners.py}{{\ttfamily example\_experiment\_for\_beginners.py}}
which runs out-of-the-box as a benchmarking Python script just as
\href{https://github.com/numbbo/coco/blob/master/code-experiments/build/python/example\_experiment2.py}{{\ttfamily example\_experiment2.py}} which also allows to run the experiment in separate batches.
}
The benchmarking scripts write experimental data into the {\ttfamily exdata} folder first,
then invoke the post-processing, {\ttfamily cocopp}, writing figures and tables into the {\ttfamily ppdata} folder (by default).
The last line of the Python script finally opens the file
{\ttfamily ppdata/index.html} in the browser as shown in
Figure~\ref{fig:browseroutput}, left, to visually investigate\del{ and analyze} the resulting data.

\lstset{style=Pythonstyle, frame=single}

\begin{figure} 
	\begin{lstlisting}[language=bash]
$ ### get and install the code
$ git clone https://github.com/numbbo/coco.git  # get coco using git
$ cd coco
$ python do.py run-python  # install Python experimental module cocoex
$ python do.py install-postprocessing install-user # install postprocessing :-)
\end{lstlisting}%\vspace{-4ex}
	
	\begin{lstlisting}[language=bash]
$ ### (optional) run an example from the shell
$ mkdir my-first-experiment
$ cd my-first-experiment
$ cp ../code-experiments/build/python/example_experiment2.py .
$ python example_experiment2.py  # run the current "default" experiment
$                                # and the post-processing
$                                # and open browser when finished
\end{lstlisting}%\vspace{-4ex}
	
% \begin{python}
  \begin{lstlisting}[language=python]
#!/usr/bin/env python
"""Python script to benchmark fmin of scipy.optimize"""
from __future__ import division  # not needed in Python 3
import cocoex, cocopp  # experimentation and post-processing modules
import scipy.optimize # to define the solver to be benchmarked

### input
suite_name = "bbob"
output_folder = "scipy-optimize-fmin"
fmin = scipy.optimize.fmin
budget_multiplier = 2  # increase to 10, 100, ...

### prepare
suite = cocoex.Suite(suite_name, "", "")
observer = cocoex.Observer(suite_name, "result_folder: " + output_folder)

### go
for problem in suite:  # this loop will take several minutes or longer
    problem.observe_with(observer)  # will generate the data for cocopp
    # restart until the problem is solved or the budget is exhausted
    while (not problem.final_target_hit and
           problem.evaluations < problem.dimension * budget_multiplier):
        fmin(problem, problem.initial_solution_proposal())
        # we assume that 'fmin' evaluates the final/returned solution

### post-process data
cocopp.main(observer.result_folder)  # re-run folders look like "...-001" etc
\end{lstlisting}\vspace{-1ex}
% \end{python}
	
\caption[Minimal benchmarking code in Python]{\label{fig:cococode}	
Shell code for \new{user} installation of \href{https://github.com/numbbo/coco}{COCO} (above)
\new{and for running a benchmarking experiment from a shell via Python (middle),}
and Python code to benchmark {\ttfamily scipy.optimize.fmin} on the \bbob{} suite (below).
}
\end{figure}

\newcommand{\underscore}{\!\_\hspace{0.05em}}
\subsection{Analysing the Results}
\label{sec:analyzing-the-results}
We present an illustrative overview of the data analysis possibilities offered by the \href{https://github.com/numbbo/coco}{COCO} platform.
The outputs for the data analysis are generated by the {\ttfamily cocopp} Python module which is at the core of the \href{https://github.com/numbbo/coco}{COCO} platform
and is installed either as shown in Figure~\ref{fig:cococode} (last line of the top box) or simply with the shell command
\noindent
\lstset{escapeinside={\%*}{*)}}
\begin{lstlisting}[language=bash]
python -m pip install cocopp
\end{lstlisting}
%
%\nnew{\footnote{%
%Since April 2020, the post-processing module can be installed }%footnote
The module can be accessed from the command line or within a Python/IPython shell or a Jupyter notebook.

\new{The post-processing is directly invoked by the scripts in Figure~\ref{fig:cococode}, except when {\ttfamily example\_experiment2.py} is run in several batches.
In the latter case, we move the root output folders \new{generated by each batch into a single folder,} say {\ttfamily EXP\underscore{}FOLDER}, and evoke the post-processing}\del{
the post-processing of data previously written into the {\ttfamily EXP\underscore{}FOLDER} folder
is evoked} 
by the shell command
\noindent
\lstset{escapeinside={\%*}{*)}}
\begin{lstlisting}[language=bash]
python -m cocopp EXP%*\!*)_FOLDER
\end{lstlisting}
or by typing

\lstset{escapeinside={\%*}{*)}}
\noindent
\begin{minipage}{\textwidth}
%\parbox{\textwidth}{
\begin{lstlisting}[language=python]
import cocopp
cocopp.main('EXP%*\!*)_FOLDER')
\end{lstlisting}
%}
\end{minipage}
in a Python/IPython shell or a Jupyter notebook.
\new{These commands take the experimental data from the {\ttfamily EXP\underscore{}FOLDER} folder and produce figures, tables and html-pages by default in the {\ttfamily ppdata} folder.}

Data from multiple experiments can be post-processed together by adding
folder names to these calls \new{where each folder represents a full experiment}.
Additionally, the \href{https://github.com/numbbo/coco}{COCO} data archive allows to access online data from previously run experiments.
This archive can be explored interactively as
shown in Figure~\ref{fig:archive}.

\begin{figure}
\begin{lstlisting}[language=python]
In [1]: import cocopp
In [2]: ars = cocopp.archives
In [3]: ars.all  # get the list of all available data sets
Out[3]:
['bbob/2009/ALPS_hornby_noiseless.tgz',
 'bbob/2009/AMALGAM_bosman_noiseless.tgz',
 [...]
 'bbob-noisy/2016/PSAaSmD-CMA-ES_Nishida_bbob-noisy.tgz',
 'test/N-II.tgz',
 'test/RS-4.zip']
In [4]: ars.bbob.find('bfgs')  # find all 'bbob' data sets
Out[4]:                        # with 'bfgs' in their name
['2009/BFGS_ros_noiseless.tgz',
 '2012/DE-BFGS_voglis_noiseless.tgz',
 '2012/PSO-BFGS_voglis_noiseless.tgz',
 '2014-others/BFGS-scipy-Baudis.tgz',
 '2014-others/L-BFGS-B-scipy-Baudis.tgz',
 '2018/BFGS-M-17.tgz',
 '2018/BFGS-P-09.tgz',
 '2018/BFGS-P-Instances.tgz',
 '2018/BFGS-P-range.tgz',
 '2018/BFGS-P-StPt.tgz',
 '2019/BFGS-scipy-2019_bbob_Varelas_Dahito.tgz',
 '2019/L-BFGS-B-scipy-2019_bbob_Varelas_Dahito.tgz']
In [5]: bfgs_data_link = ars.bbob.get_first('2018/BFGS-P-S')
  downloading [...]
In [6]: cocopp.main(bfgs_data_link)  # post-process data
Post-processing (1)
  Using:
    PATH-TO-USER-HOME/.cocopp/data-archive/bbob/2018/BFGS-P-StPt.tgz
  [...]
In [7]: cocopp.main("2009/NEWUOA! BFGS-P-StPt NELDERDOERR MCS! NIPOPaCMA "
                    "lmm RANDOMSEARCH!")  # post-process multiple data sets
Post-processing (2+)
  downloading http://coco.gforge.inria.fr/data-archive/...
	[...]
\end{lstlisting}\vspace{-1ex}
\caption[Use of archived data]{\label{fig:archive}
Example of \new{searching in and} post-processing\del{ and usage} of archived data.
}
\end{figure}

If a given name is not a unique data set substring,
\href{https://github.com/numbbo/coco}{COCO} provides the list of all matching data sets.
In this case, the user has three choices.
Either specify the name further until it is unique;
or, add an exclamation mark, like with {\ttfamily MCS!},
to retrieve the first entry of the matching list by the {\ttfamily  get\_first} method;
or, add a star (like {\ttfamily JADE*}) to retrieve all matching entries.
\new{Typically, o}nly results from the same function suite can be processed together.\footnote{\new{Exceptions of \emph{compatible} function suites exist---for example when they contain the same functions over different dimensions (which is the case for the \bbob{} and \bboblargescale{} suites) or when one is a subset of the other (like for the \bbobbiobj{} and \bbobbiobjext{} suites).}}

In the system shell, the last call to the post-processing in Figure~\ref{fig:archive} transcribes to

\noindent\begin{minipage}{\textwidth}%,basicstyle=\tiny gives wrong font
\begin{lstlisting}[language=python,basicstyle=\tt\fontsize{7pt}{8pt}\selectfont]
python -m cocopp 2009/NEWUOA! BFGS-P-StPt NELDERDOERR MCS! NIPOPaCMA lmm RANDOMSEARCH!
\end{lstlisting}
\end{minipage}

This call processes the data sets of seven (more-or-less representative) solvers benchmarked on the 24 unconstrained
continuous test functions of the \bbob test suite:
\begin{itemize}
\item {} 
NEWUOA, the NEW Unconstrained Optimization Algorithm \cite{POW2006} with the recommended number of
\(2n+1\) interpolation points in the quadratic model.
We show the results for a Scilab implementation, as benchmarked in \cite{ROS2009}.

\item {} 
The BFGS algorithm \cite{broyden1970bfgs,fletcher1970bfgs,goldfarb1970bfgs,shanno1970bfgs}, as implemented in the Python function {\ttfamily fmin\_bfgs} from the {\ttfamily scipy.optimize} module with the origin
as the initial point, named BFGS-P-StPt \cite{BFAB2018}.

\item {} 
The downhill simplex method by Nelder and Mead \cite{nelder1965dsm} with re-shaping and halfruns as presented in \cite{DFSW2009},
denoted as NELDERDOERR.

\item {} 
The multi\new{level}\del{ple} coordinate search (MCS, \cite{Huyer:1999}) by Huyer and Neumaier \cite{HN2009}.

\item {} 
Two variants of the Covariance Matrix Adaptation Evolution Strategy \cite{ho2001a}: a version with increasing
population size and negative recombination weights \cite{LOSH2012}, entitled NIPOPaCMA,
and a local meta-model assisted version benchmarked in \cite{ABH2013}, under the name lmm-CMA-ES.

\item {} 
As a baseline, a simple random search (RANDOMSEARCH) with solutions sampled uniformly at random in the
hyperbox \([-5,5]^n\) \cite{AUG2009}.

\end{itemize}

The post-processing runs for a few minutes and produces an output folder {\ttfamily ppdata}
where all visualizations, tables, HTML pages, etc. are written.
Once finished, a web browser opens and displays the results.

Depending on the number of compared data sets \new{(with cases $1, 2$, and $>\!2$)}, different visualizations will appear\new{, as detailed in the following}. \new{Note that all plots, displayed here, have been made with COCO, version 2.3.2.}

The main display in \href{https://github.com/numbbo/coco}{COCO} are runtime distributions (\del{runtime }ECDFs \new{of number of function evaluations}) to solve a given set of problems.
Extending over so-called data profiles \cite{MOR2009},
\href{https://github.com/numbbo/coco}{COCO} aggregates problems with different target precision values and displays
the runtime distributions with simulated restarts.
The default target \new{precision value}s are 51 evenly log-spaced values between \(10^{-8}\) and \(10^2\).

\begin{figure}
\centering
\raisebox{-\height}{\includegraphics[width=0.450\linewidth]{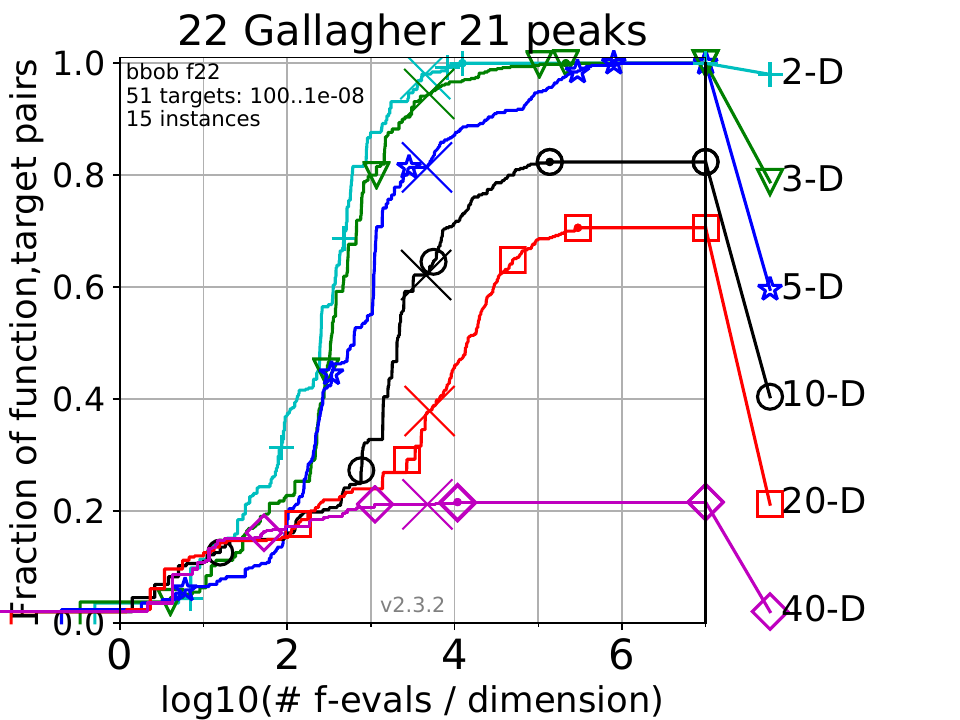}} \raisebox{-\height}{\includegraphics[width=0.450\linewidth]{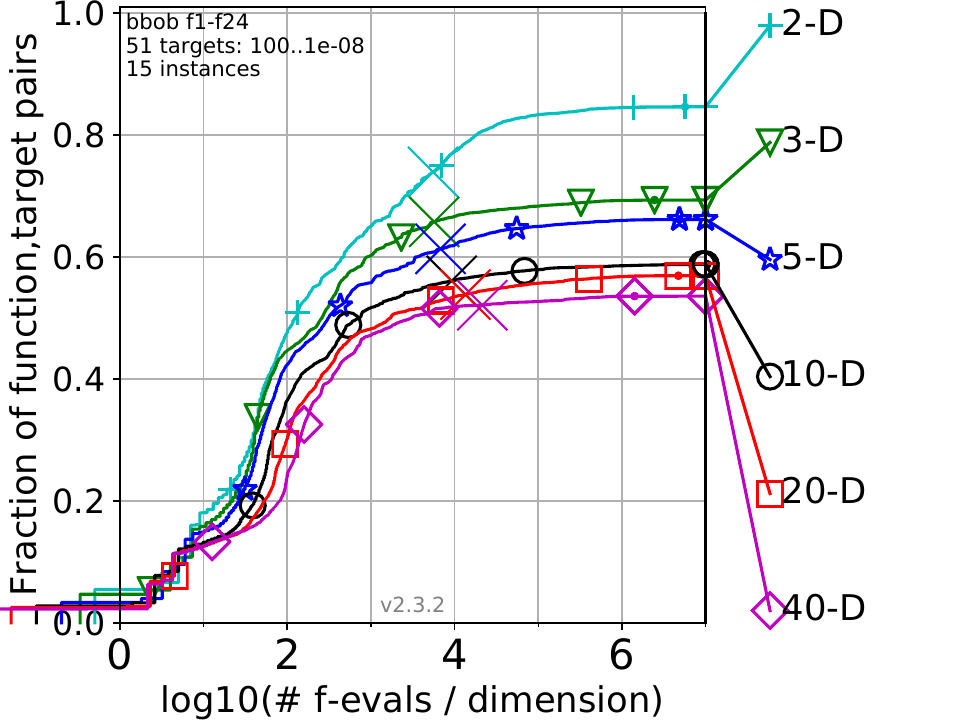}}\\[0.5em]

\raisebox{-\height}{\includegraphics[width=0.450\linewidth]{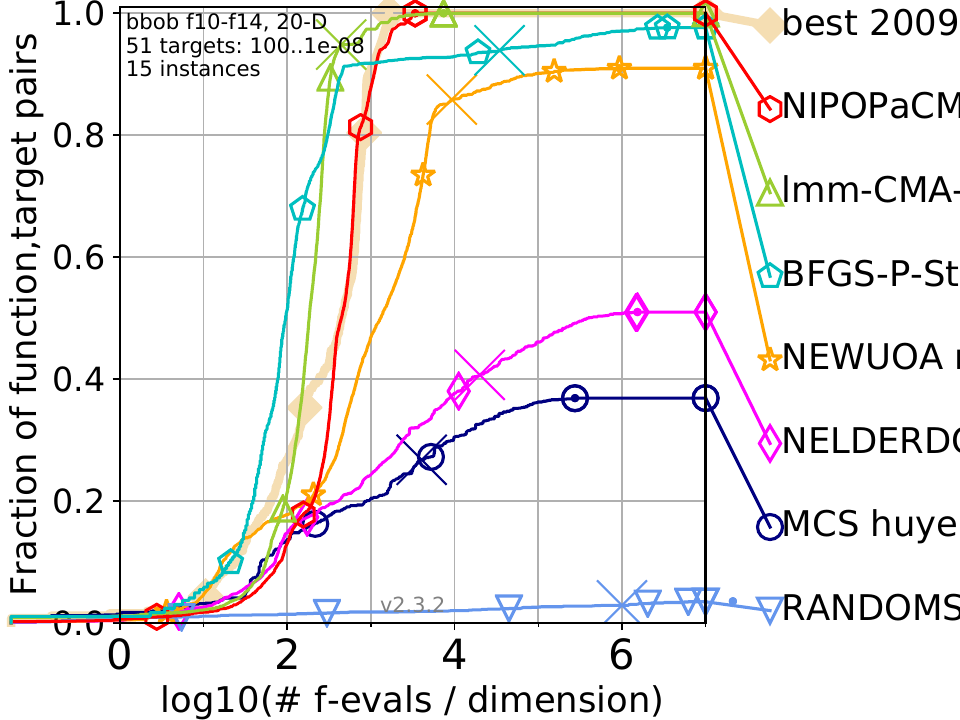}} \raisebox{-\height}{\includegraphics[width=0.450\linewidth]{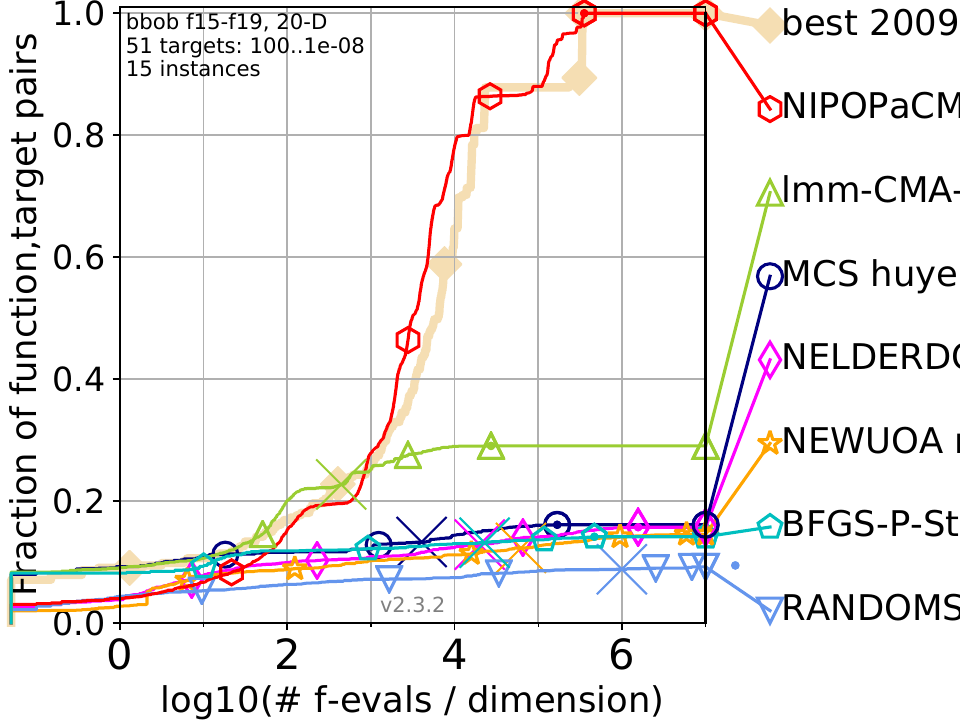}}\\[0.5em]

\raisebox{-0.5\height}{\includegraphics[width=0.450\linewidth]{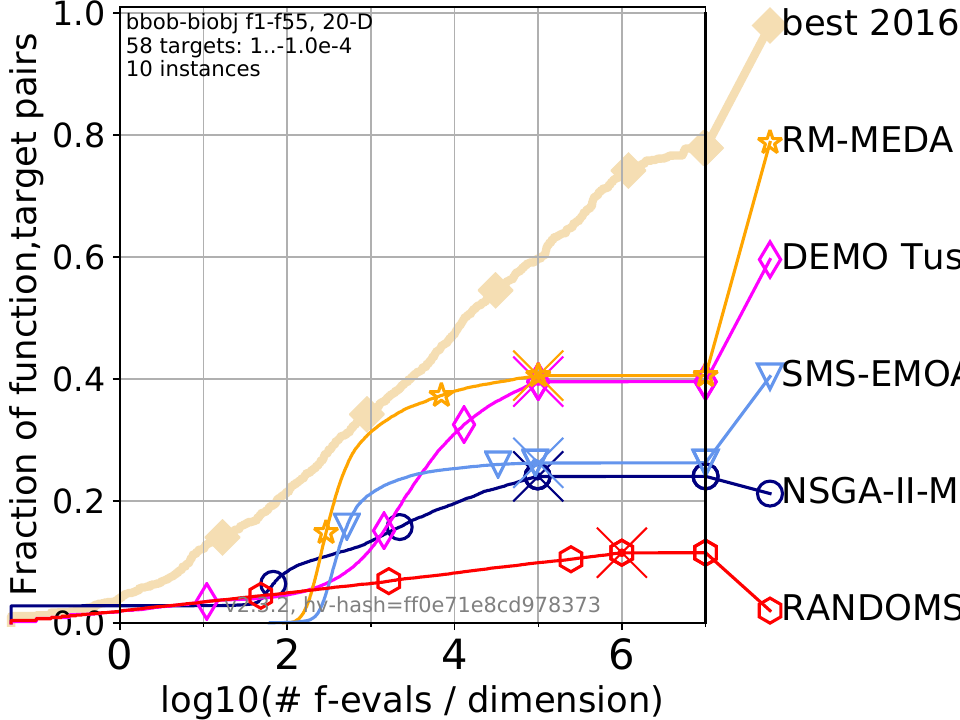}}
\raisebox{-0.5\height}{\includegraphics[width=0.450\linewidth]{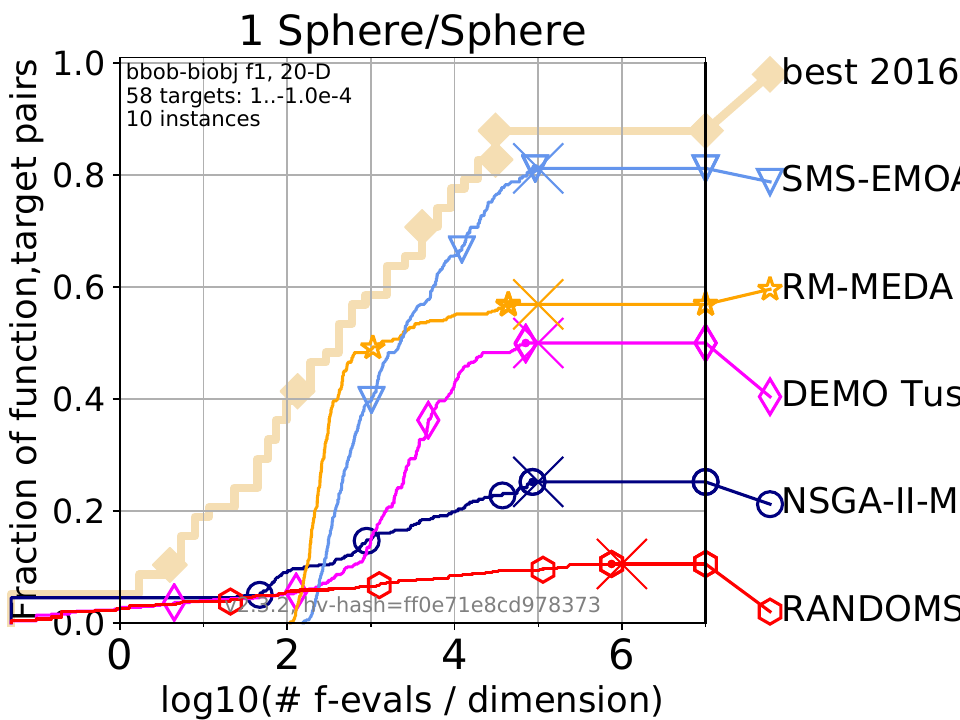}}
\caption{\label{fig:exampleECDFs}
Examples of runtime ECDF plots in \href{https://github.com/numbbo/coco}{COCO}.
First row: plots for a single solver (here: BFGS-P-StPt) in\del{
six} dimensions between 2 and 40, \new{to the left}
on the multi-modal function $f_{22}$\del{ (left)} and \new{to the right} aggregated over all 24
\bbob{} functions\del{ (right)}.
Second row:
plots for seven solvers in dimension 20, aggregated
over all problems from the highly ill-conditioned function group (left) and the
multi-modal function group with global structure (right).
Third row:
plots for five multiobjective solvers \new{in 20-D}
on the \bbobbiobj{}
test suite with data aggregated over all 55 functions\del{ in 20-D} (left)
and for the double-sphere problem \bbobbiobj{} $F_1$\del{ in 20-D} (right).
Long algorithm names are cut for readability.
}
\end{figure}
Figure~\ref{fig:exampleECDFs} shows examples of runtime ECDFs from single functions, aggregated over
function groups, and aggregated over all functions of a test suite.
Plots of the first two rows were generated with the above {\ttfamily cocopp} calls. The last
row shows results for five well-known multiobjective algorithms on the \bbobbiobj
suite, invoked by the command

\lstset{escapeinside={\%*}{*)}}
\begin{lstlisting}{language=bash,showspaces=True}
python -m cocopp NSGA-II! DEMO RM-MEDA SMS-EMOA! bbob-biobj/2016/RANDOMSEARCH%*\!*)-5
\end{lstlisting}
The multiobjective algorithms are NSGA-II \cite{dapm2002a,AUG2016ns}, DEMO \cite{robivc2005differential,TUS2016}, RM-MEDA \cite{zhang2008rm,AUG2016rm}, SMS-EMOA-DE \cite{bne2007a,AUG2016sms},
and a uniform random search within the hyperbox
\([-5,5]^n\) \cite{AUG2016rnd}.

\begin{figure}
\centering
\raisebox{-\height}{\includegraphics[width=0.450\linewidth]{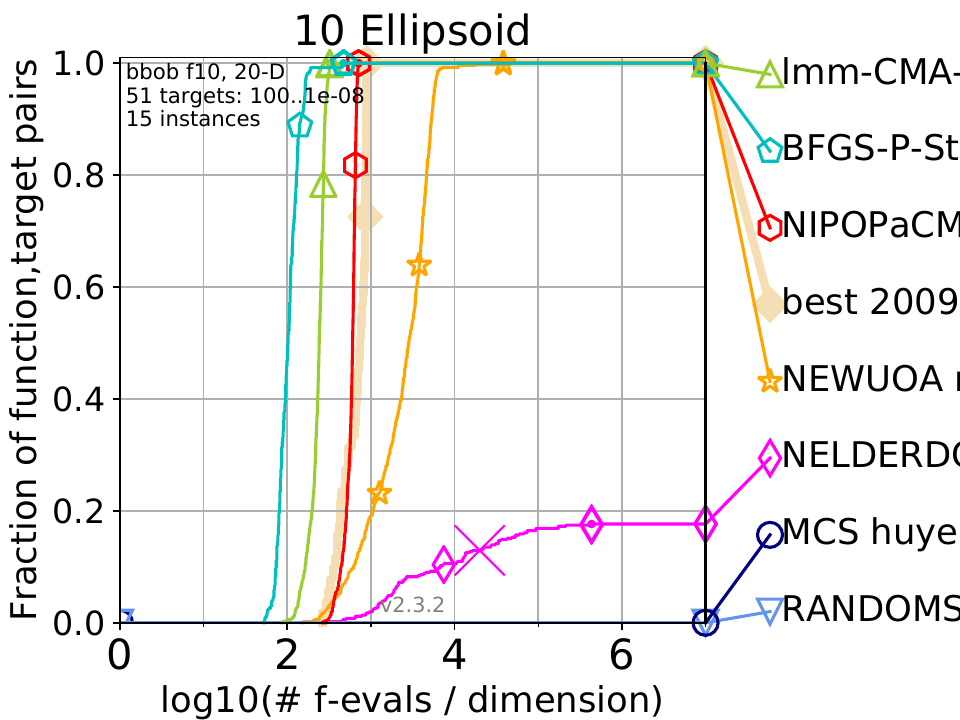}} \raisebox{-\height}{\includegraphics[width=0.450\linewidth]{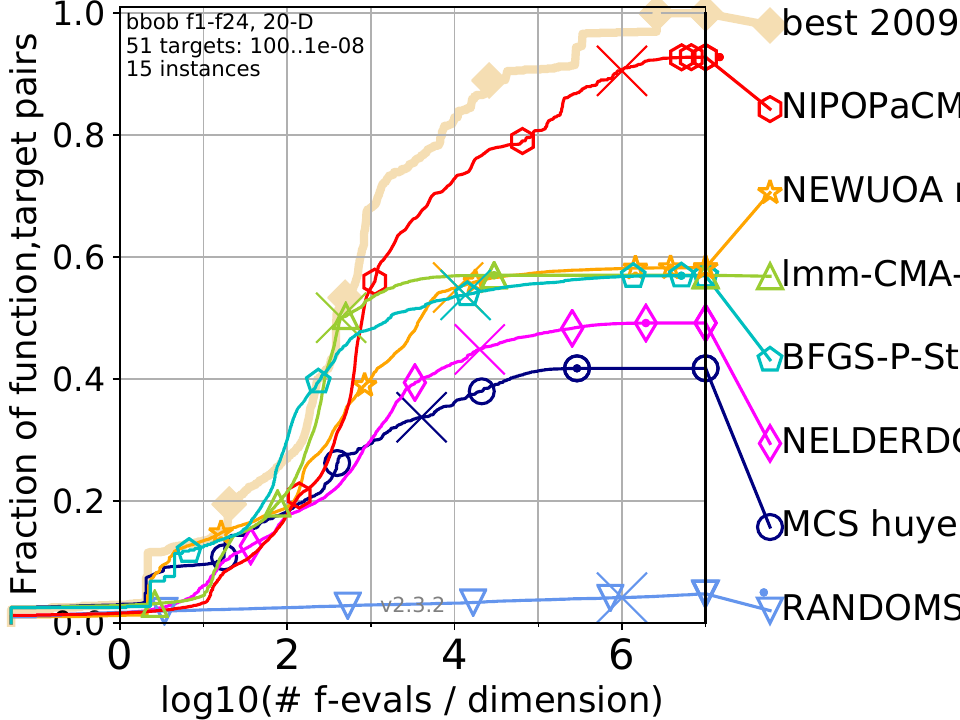}}\\[0.5em]

\raisebox{-\height}{\includegraphics[width=0.450\linewidth]{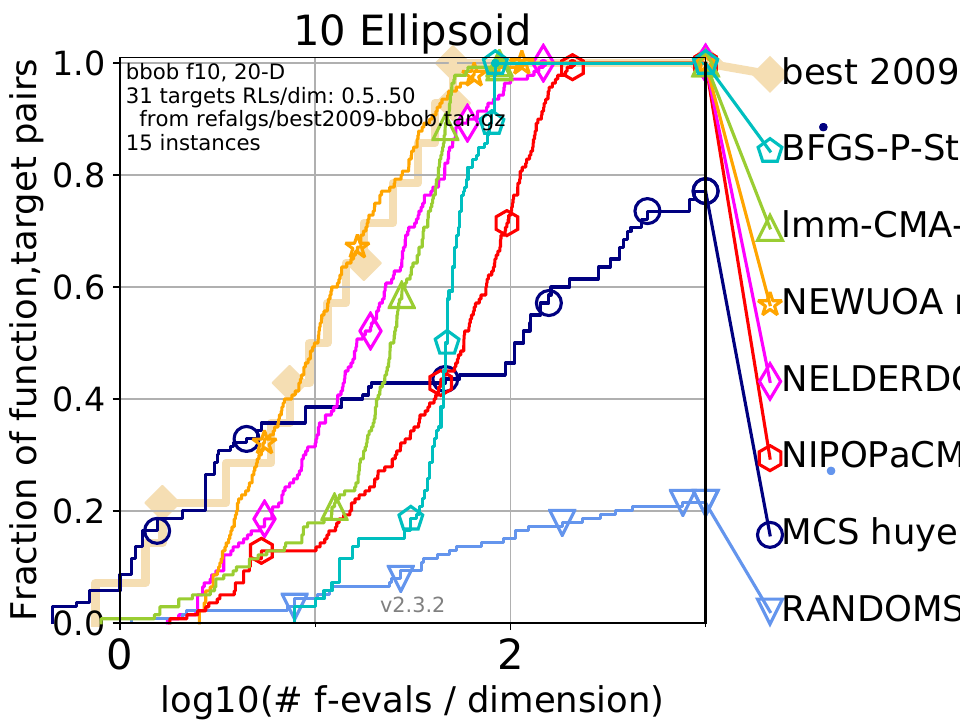}} \raisebox{-\height}{\includegraphics[width=0.450\linewidth]{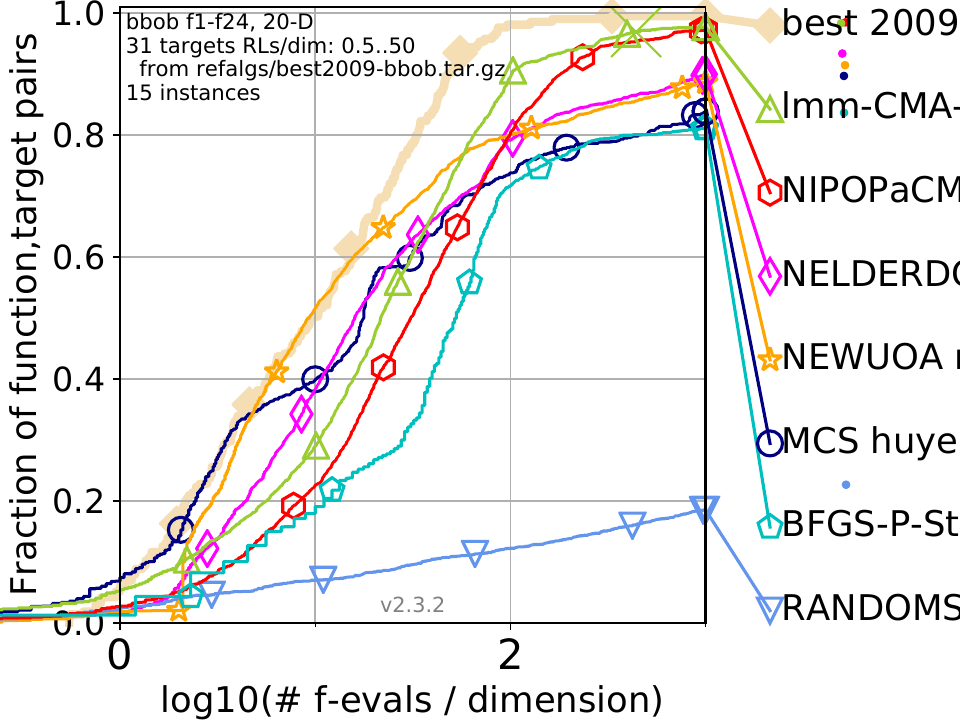}}
\caption{\label{fig:exampleECDFtwo}
Examples of runtime ECDF plots in \href{https://github.com/numbbo/coco}{COCO}
and the impact of the target choice.
First row:
plots for seven solvers in dimension 20 on a single function (left) and aggregated over all
24 \bbob
functions (right) for the 51 default target \new{precision}s, equidistantly log-spaced
between 100 and \(10^{-8}\).
Second row: as first row, but using budget-based targets
with the target-wise best result from the BBOB-2009 workshop as reference
and with budgets between 0.5 and 50 times dimension function evaluations.
}
\end{figure}

To \new{demonstrate}\del{show} the influence of the target choice, Figure~\ref{fig:exampleECDFtwo} shows runtime \new{distributions}\del{ECDF plots} for the same\del{ seven single-objective} solvers \new{as in}\del{of} Figure~\ref{fig:exampleECDFs} for \new{(i)} the 51 default target values (first row)
and \new{(ii)} for 31 budget-based target values\new{, see Section~\ref{sec:runtime-and-target-values},} with budgets up to fifty times dimension
evaluations. \new{Budget-based targets are} invoked by \new{simply adding the optional argument} ``{\ttfamily --expensive}'' \new{to the {\ttfamily cocopp} call.}\del{option of \href{https://github.com/numbbo/coco}{COCO}.}
Results are shown in dimension 20 on the function $f_{10}$ (left column) and aggregated over all 24
\bbob functions (right column).
The absolute targets (\new{first row}\del{above}) do not reveal the initially superior speed of MCS, NEWUOA, or Nelder-Mead.
The \new{budget-based}\del{relative} targets (\new{second row}\del{below}) do not reveal that the (disturbed) ellipsoid $f_{10}$ is not even closely solved by MCS or Nelder-Mead.
\par
\begin{figure}
\centering
\raisebox{-\height}{\includegraphics[width=0.450\linewidth]{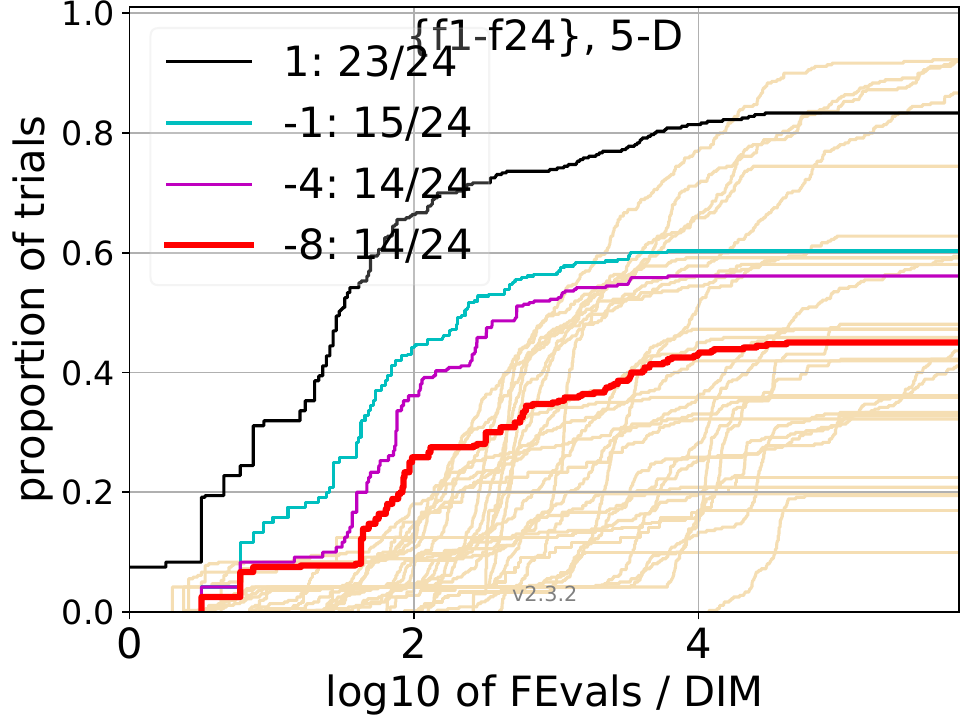}} \raisebox{-\height}{\includegraphics[width=0.450\linewidth]{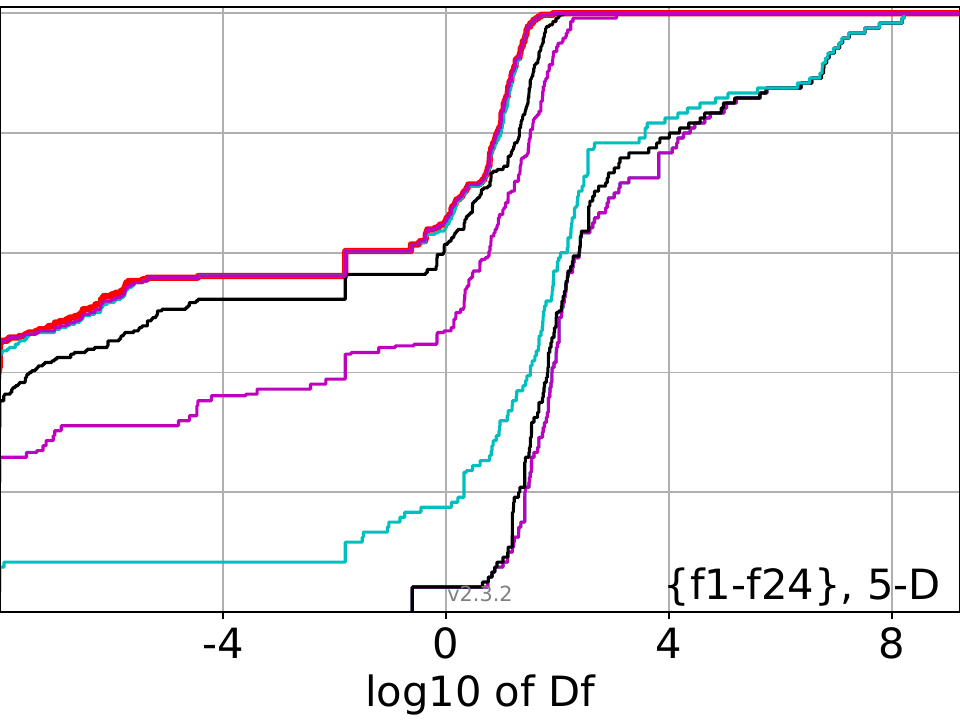}}
\caption{\label{fig:exampleECDFthree}
ECDFs of runtimes to \new{reach} a single target precision\del{ for different targets} (left) and
ECDFs of precisions reached for different budgets (right) \new{for BFGS-P-StPt}.
Left: the target \new{precision}s for each graph \new{are} from left to right, respectively,
\(10^1\), \(10^{-1}\), \(10^{-4}\), and \(10^{-8}\),
and the legend provides the number of functions solved for each \del{target}\new{precision}.
Right: the budgets for each\del{ of the seven} graph from right to left are, respectively,
\(0.5\), \(1.2\), \(3\), \(10\), \(100\), \(1000\), \new{and \(10\,000\)}
times dimension and the maximal budget
(thick red) indicating the final distribution of \new{precisions}\del{\(f\)-} values.
The left most point of each precision graph \new{on the right,}
representing precision \(10^{-8}\),
coincides with the (thick red) runtime graph for \new{precision}\del{target} \(10^{-8}\)
at the respective budgets \new{and proportion of trials} to the left.
The same holds analogously for all shown target \new{precision}s.
}
\end{figure}

When only a single solver is post-processed, \href{https://github.com/numbbo/coco}{COCO}
produces also single-target\new{-precision} runtime ECDFs \new{(akin to data profiles})
for four different target precision values,
along with ECDFs of the precisions
attained within a given budget, as shown in Figure~\ref{fig:exampleECDFthree}.
Obtained precision ECDFs extend data profile graphs naturally to the right (see also \new{the} caption).
\par
\begin{figure}
\centering
\raisebox{-0.5\height}{\includegraphics[width=0.450\linewidth]{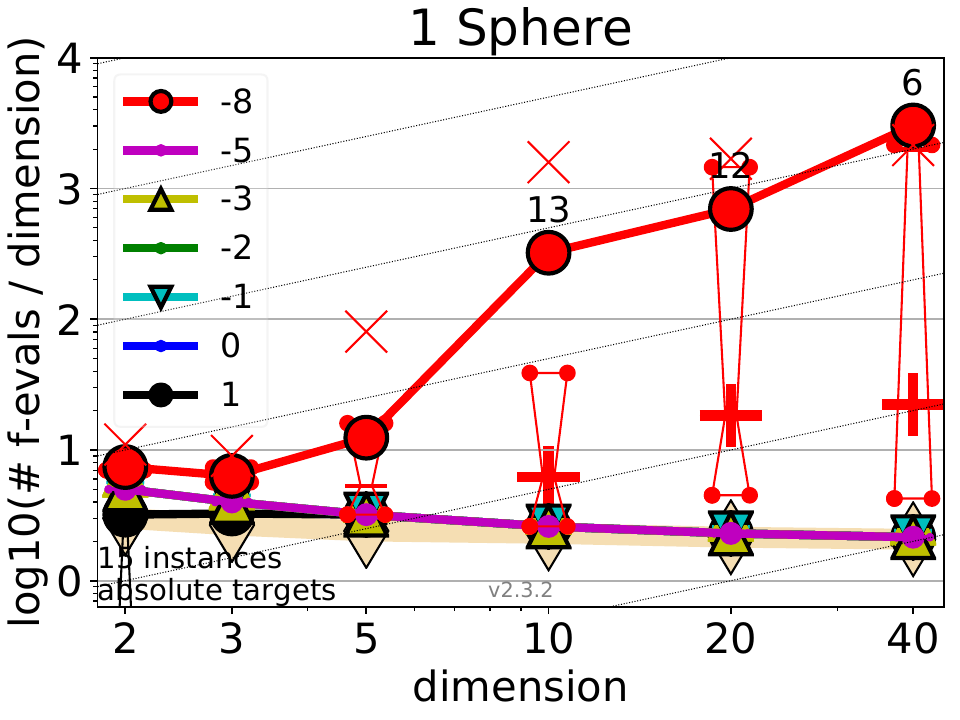}} \raisebox{-0.5\height}{\includegraphics[width=0.450\linewidth]{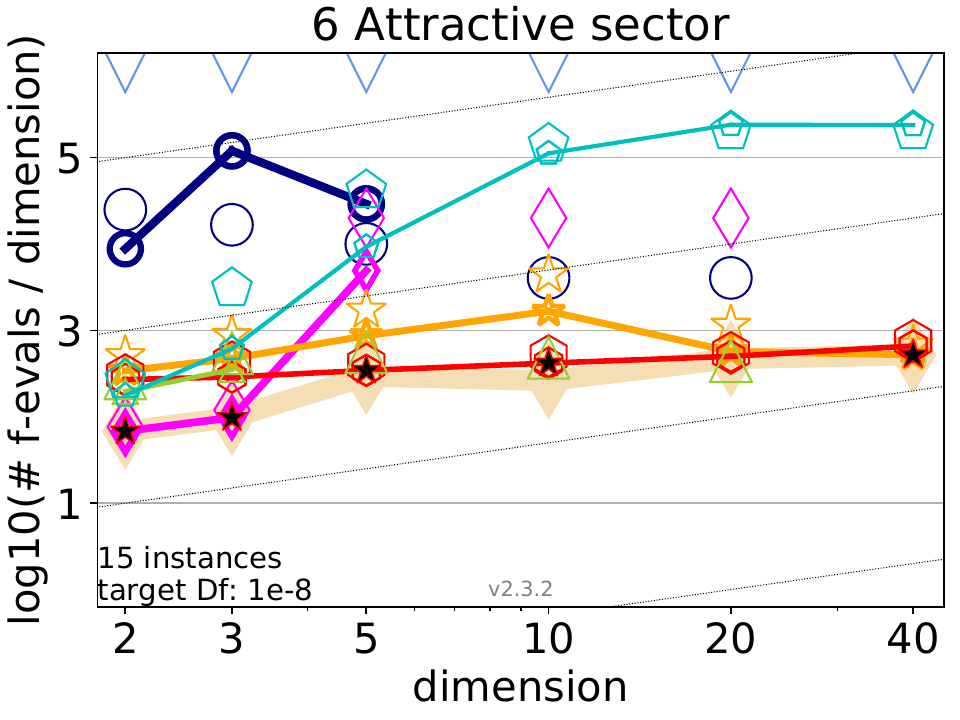}}
\caption{\label{fig:scalingplot}
Scaling \new{with the dimension}\del{plots} \new{of BFGS-P-StPt} on the sphere function \new{for different target precisions} (left) and\del{
for a single target \new{precision} but} of different solvers on the attractive sector function \new{for target precision $10^{-8}$} (right\new{, the algorithms are distinguished by the same colors and markers as in Figure~\ref{fig:exampleECDFtwo}}).
Shown are\del{ the} average runtimes divided by the problem dimension (in log\new{10}-scale,
only the exponent is annotated) to reach a given target,
plotted against dimension.
The corresponding\del{ (relative)} target \new{precision}s are either mentioned
in the legend with the number \(i\) indicating target \new{precision} \(10^{i}\) (left plot)
or at the bottom left of the figure as ``target Df'' (here \(10^{-8}\) in the right plot).
}
\end{figure}

Additionally, \href{https://github.com/numbbo/coco}{COCO}
generates scaling plots of the average runtime (ERT) over dimension.
Figure~\ref{fig:scalingplot}
shows the scaling of \new{BFGS-P-StPt}\del{for a single solver} for several target \new{precisions} (left plot)
and the scaling of multiple solvers for a single target \new{precision} (right plot).
These bring a direct visual aid to investigating
how the performance of a solver scales with the problem dimension.
\par
\begin{figure}
\centering
\raisebox{-\height}{\includegraphics[width=0.250\linewidth]{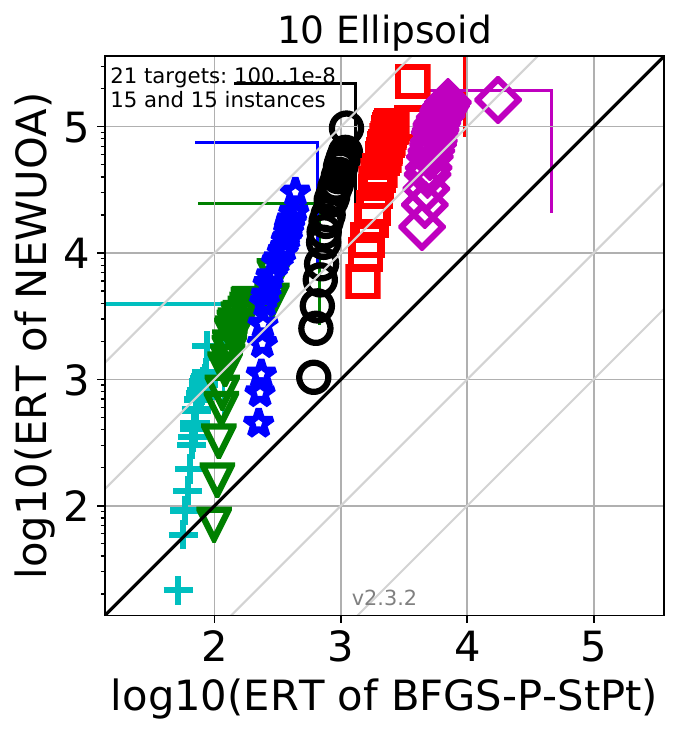}} \raisebox{-\height}{\includegraphics[width=0.730\linewidth]{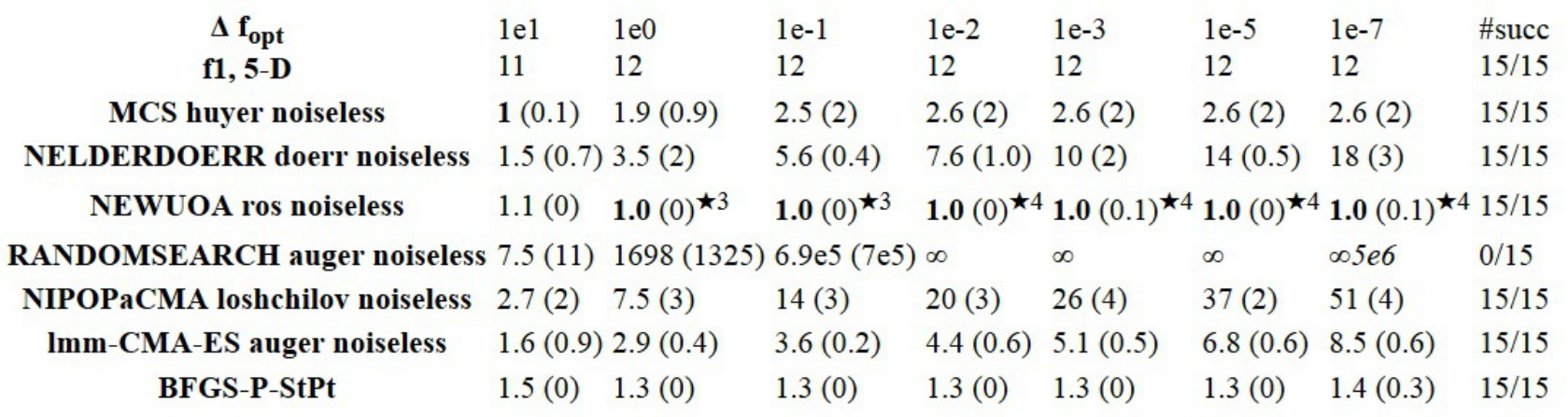}}
\caption{\label{fig:scatterplot}
Examples of the \href{https://github.com/numbbo/coco}{COCO} scatter plots (left) and the platform's tabular output (right).
The scatter plot shows,
for 21 target \new{precision value}s between 100 and \(10^{-8}\) per dimension,
the average runtime (ERT) in log10 of the number of function evaluations
for the two solvers BFGS-P-St-Pt and NEWUOA.
The thin \new{colored axis-aligned line segments}\del{rectangular line} indicate
the maximal budgets
associated with the x- and y-axis.
The table shows ERT ratios for the target \new{precision}s given in the first row.
The ratios are computed by dividing ERT with the best ERT from 31 solvers from
\href{http://coco.gforge.inria.fr/doku.php?id=bbob-2009-results}{BBOB-2009}
(given in the second line).
The dispersion measure given in brackets is the semi-interdecile range
(half the difference between the 10 and 90\%-tile)
of bootstrapped runtimes.
If the last target was never reached, the median number of conducted function evaluations is given in italics.
The last column (\#succ) contains the number of
trials that reached the (final) target \(f_{\text opt}+ 10^{-8}\).
Entries succeeded by a star are statistically significantly better
(according to the rank-sum test) when compared to all other solvers of the table,
with \(p = 0.05\) or \(p = 10^{-k}\)
when a number \(k\) follows the star, with Bonferroni
correction by the number of functions (24). Best results are printed in bold.
}
\end{figure}

When only two data sets are compared, \href{https://github.com/numbbo/coco}{COCO} also produces for each function
a scatter plot of ERT values for 21 targets and all dimensions, as shown in
Figure~\ref{fig:scatterplot}.
Performance data are also available in a tabular format, an example thereof is
shown on the right-hand side in Figure~\ref{fig:scatterplot}.
\par
\begin{figure}
\centering
\noindent{\hspace*{\fill}\includegraphics[width=0.800\linewidth]{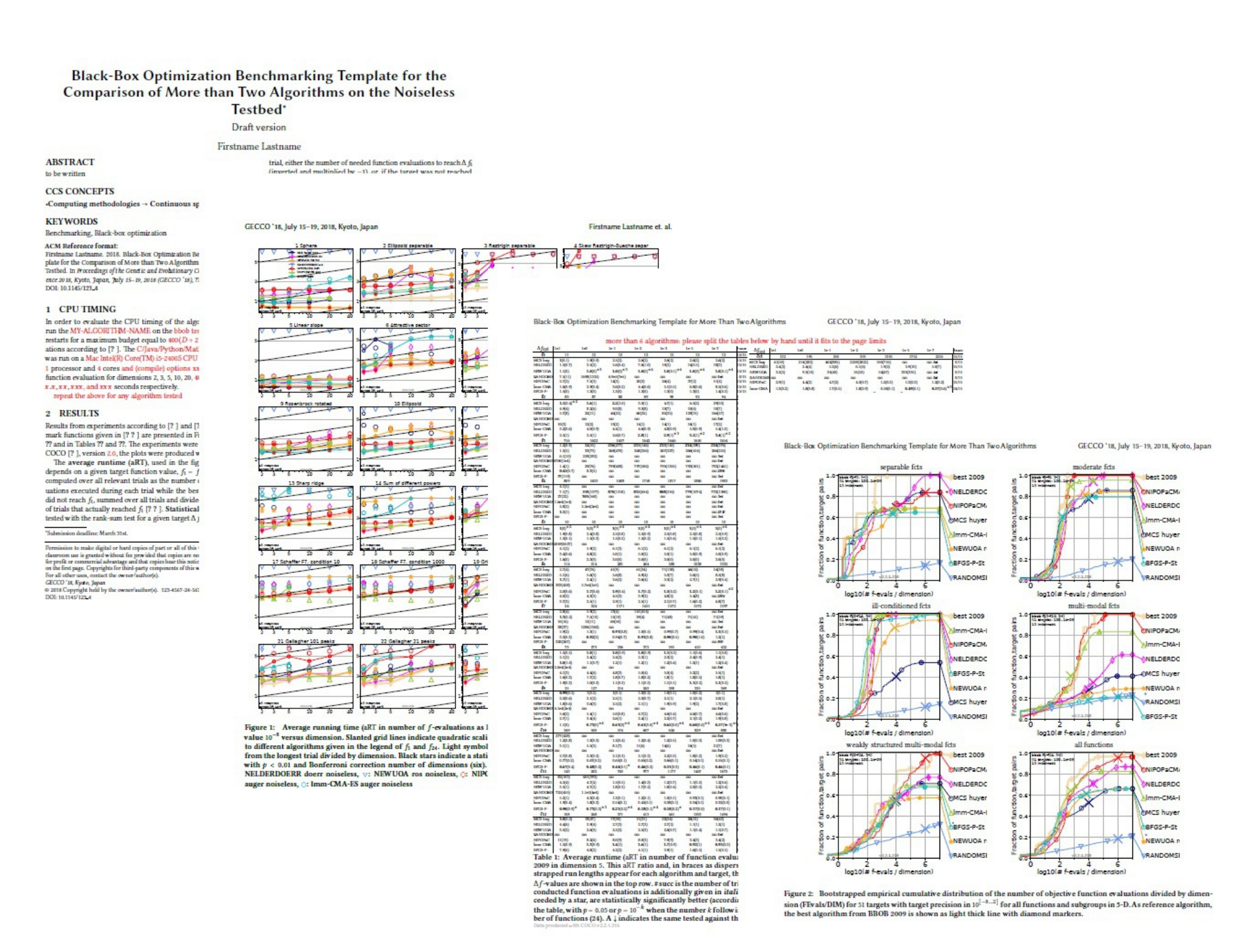}\hspace*{\fill}}
\caption{\label{fig:template}
\new{Compiled LaTeX-}template for producing ACM-compliant papers from \href{https://github.com/numbbo/coco}{COCO} data.
}
\end{figure}

Besides the browser output, \href{https://github.com/numbbo/coco}{COCO}
also provides ACM-compliant LaTeX templates with already
included main performance displays, which facilitates the publication
of benchmarking results, see Figure~\ref{fig:template}.
The showcased plots are non-exhaustive and \href{https://github.com/numbbo/coco}{COCO} also provides more
detailed descriptions.

\section{Usage Statistics}
\label{\detokenize{index:usage-statistics}}
Since 2009, the \href{https://github.com/numbbo/coco}{COCO} software framework has been used to facilitate submissions to the workshops on Black-Box Optimization Benchmarking (BBOB) at the ACM GECCO conference. With time, the platform has also been more and more used outside the scope of this workshop series. Table 1 summarizes some numbers of visibility and user base, including the number of citations to the documentation.

\newcommand{\tabspace}{\hspace{11em}}
\begin{table}[h]
\caption{\textbf{Visibility of COCO.}\del{
  \emph{Papers in Google Scholar} refers to the search phrase {\emph ``comparing continuous optimizers''
  OR ``black-box optimization benchmarking (BBOB)''} and
  the \emph{COCO documentation} includes \cite{HAN2016co}, \cite{HAN2009fun}, \cite{FIN2010}, \cite{HAN2009noi}, \cite{HAN2009}, \cite{HAN2010ex}, \cite{HAN2016ex}, \cite{HAN2016perf}, \cite{BRO2016}, and \cite{BRO2019}.}
All citations as of \del{March 19, 2019}\new{November 19, 2019}, in Google Scholar.}
\noindent\begin{tabular}{p{85mm} p{40mm} r}
\hline

\parbox{0em}{\mbox{Data sets online}}%
\tabspace \bbob{} suite & 227 \\%\newline
\tabspace \bbobnoisy{} suite & \,~45 \\ %\newline
\tabspace \bbobbiobj{} suite & \,~32 \\%\newline
\tabspace \bboblargescale{} suite & \,~11 \\ %\newline
\tabspace \nnew{\bbobmixint{} suite} & \,\,~~\nnew{4}
% \tabspace \bbobbiobjmixint{} suite & \,\,~~4
% \todo{add biobj-mixint?}
\\
\hline
BBOB workshop papers using \href{https://github.com/numbbo/coco}{COCO}
&
143
\\
\hline
Unique authors on the workshop papers
&
109 from 28 countries
\\
\hline
Papers in Google Scholar
\new{found with the search phrase \emph{``comparing continuous optimizers''
  OR ``black-box optimization benchmarking (BBOB)''}}
&
~\!\!\new{559}\del{489}
\\
\hline
Citations to the \href{https://github.com/numbbo/coco}{COCO} documentation
\new{including \cite{HAN2016co} \cite{HAN2009fun} \cite{FIN2010} \cite{HAN2009noi} \cite{HAN2009} \cite{HAN2010ex} \cite{HAN2016ex} \cite{HAN2016perf} \cite{BRO2016} and \cite{BRO2019}}
&
~\!\!\new{1,455}\del{1,467}
\\
\hline\end{tabular}

\end{table}

\section{Extensions under Development}
\label{sec:extensions}
The \href{https://github.com/numbbo/coco}{COCO} software framework is under continuous development.
In this section, we briefly discuss features which have been thoroughly specified for inclusion or are already under advanced development.
\begin{itemize}
\item {} 
Interface for the implementation of new suites under Python.
In the current release, new suites can
be implemented and (seamlessly) integrated only on the C level.%
\footnote{See \href{https://github.com/numbbo/coco/blob/master/howtos/create-a-suite-howto.md}{https://github.com/numbbo/coco/blob/master/howtos/create-a-suite-howto.md}.}
The new development shall allow to integrate new benchmark suites also written
with Python, which will reduce the necessary development time considerably
and make rapid prototyping of benchmark suites possible.
\item {} 
\new{Interface for enabling external evaluation of solutions based on socket communication. The interface where an external evaluator (the server) responds to requests for evaluations by COCO  (the client) can be especially useful for supporting benchmarking on real-world problems that rely on particular software for solution evaluation.}
\item {} 
Benchmark suites for
\begin{itemize}
\item {} 
constrained problems\del{(see also Section~\ref{sec:test-suites})},

\item {} 
multiobjective problems with 3 objectives implemented in Python,

\item \new{two types of real-world problems in games, the problem of forming a deck for the TopTrumps card game and the problem of generating a level for the Super Mario Bros.\ game \cite{VOL2019} (both in single- and bi-objective variants).}

\end{itemize}

\item {} 
Use of recommendations, in order to address a (usually) small bias in the performance evaluation of noisy functions.
Recommendations represent the current return value of the solver.
The performance quality indicator is based on a short history of recommendations.

\item {} 
Re\new{writing}\del{coding} of the post-processing with interactive figures.

\end{itemize}

These features have not been released yet and their description and implementation may still undergo some relevant changes before their release.

\section{Summary and Discussion}
\label{\detokenize{index:id129}}\label{\detokenize{index:summary-and-discussion}}
We have presented the open source  \new{zero-order black-box optimization} benchmarking platform \href{https://github.com/numbbo/coco}{COCO} that allows to
benchmark numerical optimization algorithms automatically.
The platform is composed
of an interface to several languages (currently  C/C++, Java,
Matlab/Octave and Python) where the solvers can be plugged and
run on a set of test functions. The ensuing collected data
are then post-processed by a Python module and different graphs and
tables are generated. All collected benchmarking data are open
access, allowing to more easily reproduce and in particular seamlessly compare results.
As of \del{March 2019}\new{November 2019}, more than \del{250}\new{300} datasets are available.
The platform also supports the implementation of new test suites.

The benchmarking methodology implemented within the platform is original
in several key aspects:
\begin{itemize}
\item {} 
Each test function comes in several instances that typically differ by having different (pseudo-randomly sampled) optima, shifts of function value, and rotations.
The underlying assumption when analysing the data is that different instances of the same test function have similar difficulties. This notion of function instances allows to compare deterministic and stochastic solvers in a single framework and makes it harder for a solver to exploit specific function instance properties (like a specific position of the optimum).

\item {} 
All test functions are scalable with respect to the input dimension.

\item {} 
The \bbob test suite for unconstrained optimization tries to reflect difficulties encountered in reality. A well-balanced set of difficulties and a scalable testbed is especially important when performance is aggregated, for example, through runtime ECDFs or data profiles.

\item {} 
We never aggregate results over dimension because dimension is a \emph{known}
input to the solver which can and should be used when deciding
which solver to apply on a (real-world) problem.
We may aggregate, however, over many target values.

\item {} 
The benchmarking methodology generalizes to multiobjective problems by using a quality indicator
which maps all so-far evaluated solutions to a single value and
defining targets for this indicator value.
Currently, only benchmarking of bi-objective problems is supported.

\end{itemize}

As a final remark, we want to emphasize the importance of carefully
scrutinizing the test functions used in aggregated results,
for example,
in empirical runtime distributions and data or performance profiles.
Unbalanced test suites (for example, primarily low-dimensional, separable, ...)
and the common approach to aggregate over all
functions from a suite
can lead to strong biases.
This may disconnect solvers that perform well on test suites from those
that perform well on real-world problems and hence seriously misguide research
efforts in numerical optimization.

If we aim towards providing software that can address real-world difficulties, we should
(i) include in our test suites
mainly challenging, yet solvable problems and
(ii) ensure that aggregated performance measures do not over-emphasize
results on unimportant problem classes---like easy-to-solve problems.

\section*{Acknowledgments}
The authors would like to thank\del{ Raymond Ros,} Steffen Finck, Marc Schoenauer,
Petr Po\v{s}ik and Dejan Tu\v{s}ar for their many invaluable contributions to this work.

The authors also acknowledge support by the grant ANR-12-MONU-0009 (NumBBO)
of the French National Research Agency. This work was furthermore supported by a public grant as
part of the Investissement d'avenir project, reference ANR-11-LABX-0056-LMH, LabEx LMH, in a joint call with Gaspard Monge Program for optimization, operations research and their interactions with data sciences.

Tea Tu\v{s}ar acknowledges the financial support from the Slovenian Research Agency \nnew{(research project No.\ Z2-8177 and research program No.\ P2-0209) and the European Commission's Horizon 2020 research and innovation program (grant agreement No.\ 692286)}.

\bibliographystyle{tfs}
\bibliography{bbob,all}

\begin{thebibliography}{10}
\providecommand{\MR}{\relax\unskip\space MR }
\providecommand{\url}[1]{\normalfont{#1}}
\providecommand{\urlprefix}{Available at }

\bibitem{audet2017derivative}
C. Audet and W. Hare, \emph{Derivative-free and blackbox optimization},
  Springer, 2017.

\bibitem{ABH2013}
A. Auger, D. Brockhoff, and N. Hansen, \emph{{Benchmarking the Local Metamodel
  CMA-ES on the Noiseless BBOB'2013 Test Bed}}, in \emph{GECCO (Companion)
  workshop on Black-Box Optimization Benchmarking (BBOB'2013)}. ACM, 2013, pp.
  1225--1232.

\bibitem{AUG2016ns}
A. Auger, D. Brockhoff, N. Hansen, D. Tu{\v{s}}ar, T. Tu{\v{s}}ar, and T.
  Wagner, \emph{{Benchmarking MATLAB's gamultiobj (NSGA-II) on the Bi-objective
  BBOB-2016 Test Suite}}, in \emph{GECCO (Companion) workshop on Black-Box
  Optimization Benchmarking (BBOB'2016)}. ACM, 2016, pp. 1233--1239.

\bibitem{AUG2016rm}
A. Auger, D. Brockhoff, N. Hansen, D. Tu{\v{s}}ar, T. Tu{\v{s}}ar, and T.
  Wagner, \emph{{Benchmarking RM-MEDA on the Bi-objective BBOB-2016 Test
  Suite}}, in \emph{GECCO (Companion) workshop on Black-Box Optimization
  Benchmarking (BBOB'2016)}. ACM, 2016, pp. 1241--1247.

\bibitem{AUG2016rnd}
A. Auger, D. Brockhoff, N. Hansen, D. Tu{\v{s}}ar, T. Tu{\v{s}}ar, and T.
  Wagner, \emph{{The Impact of Search Volume on the Performance of RANDOMSEARCH
  on the Bi-objective BBOB-2016 Test Suite}}, in \emph{GECCO (Companion)
  workshop on Black-Box Optimization Benchmarking (BBOB'2016)}. ACM, 2016, pp.
  1257--1264.

\bibitem{AUG2016sms}
A. Auger, D. Brockhoff, N. Hansen, D. Tu{\v{s}}ar, T. Tu{\v{s}}ar, and T.
  Wagner, \emph{{The Impact of Variation Operators on the Performance of
  SMS-EMOA on the Bi-objective BBOB-2016 Test Suite}}, in \emph{GECCO
  (Companion) workshop on Black-Box Optimization Benchmarking (BBOB'2016)}.
  ACM, 2016, pp. 1225--1232.

\bibitem{AUG2005}
A. Auger and N. Hansen, \emph{Performance Evaluation of an Advanced Local
  Search Evolutionary Algorithm}, in \emph{Proceedings of the {IEEE} Congress
  on Evolutionary Computation ({CEC} 2005)}. 2005, pp. 1777--1784.

\bibitem{AUG2009}
A. Auger and R. Ros, \emph{Benchmarking the pure random search on the
  {BBOB}-2009 testbed}, in Rothlauf  \cite{DBLP:conf/gecco/2009c}, 2009, pp.
  2479--2484.

\bibitem{BAR1995}
R.S. Barr, B.L. Golden, J.P. Kelly, M.G. Resende, and W.R. Stewart,
  \emph{Designing and reporting on computational experiments with heuristic
  methods}, Journal of heuristics 1 (1995), pp. 9--32.

\bibitem{BHL2017}
V. Beiranvand, W. Hare, and Y. Lucet, \emph{Best practices for comparing
  optimization algorithms}, Optimization and Engineering 18 (2017), pp.
  815--848.

\bibitem{bne2007a}
N. Beume, B. Naujoks, and M. Emmerich, \emph{{SMS-EMOA: Multiobjective
  Selection Based on Dominated Hypervolume}}, European Journal of Operational
  Research 181 (2007), pp. 1653--1669.

\bibitem{BFAB2018}
A. Blelly, M. Felipe-Gomes, A. Auger, and D. Brockhoff, \emph{Stopping
  Criteria, Initialization, and Implementations of BFGS and their Effect on the
  BBOB Test Suite}, in \emph{GECCO (Companion) workshop on Black-Box
  Optimization Benchmarking (BBOB'2009)}. 2018.

\bibitem{BLTZ2003}
S. Bleuler, M. Laumanns, L. Thiele, and E. Zitzler, \emph{{PISA---A Platform
  and Programming Language Independent Interface for Search Algorithms}}, in
  \emph{Conference on Evolutionary Multi-Criterion Optimization {(EMO~2003)}},
  C.M. Fonseca, \emph{et~al.}, eds., LNCS Vol. 2632, Berlin. Springer, 2003,
  pp. 494--508.

\bibitem{BOS2018}
J. Bossek, \emph{Performance assessment of multi-objective evolutionary
  algorithms with the R package ecr}, in \emph{Companion Proceedings of the
  Genetic and Evolutionary Computation Conference (GECCO~2018)}. ACM, 2018, pp.
  1350--1356.

\bibitem{BRO2015}
D. Brockhoff, T.D. Tran, and N. Hansen, \emph{{Benchmarking Numerical
  Multiobjective Optimizers Revisited}}, in \emph{Genetic and Evolutionary
  Computation Conference (GECCO 2015)}. ACM, 2015, pp. 639--646.

\bibitem{BRO2019}
D. Brockhoff, T. Tu{\v s}ar, A. Auger, and N. Hansen, \emph{Using
  well-understood single-objective functions in multiobjective black-box
  optimization test suites}, ArXiv e-prints
  \href{https://arxiv.org/abs/1604.00359v3}{arXiv:1604.00359v3} (2019). Last
  updated January 4, 2019.

\bibitem{BRO2016}
D. Brockhoff, T. Tu{\v s}ar, D. Tu{\v s}ar, T. Wagner, N. Hansen, and A. Auger,
  \emph{Biobjective performance assessment with the {COCO} platform}, ArXiv
  e-prints \href{https://arxiv.org/abs/1605.01746}{arXiv:1605.01746} (2016).

\bibitem{broyden1970bfgs}
C.G. Broyden, \emph{The convergence of a class of double-rank minimization
  algorithms}, Journal of the Institute of Mathematics and Its Applications 6
  (1970), pp. 76--90.

\bibitem{bubeck2014convex}
S. Bubeck, \emph{Convex optimization: Algorithms and complexity} (2014).

\bibitem{BUS2014}
M.R. Bussieck, S.P. Dirkse, and S. Vigerske, \emph{Paver 2.0: an open source
  environment for automated performance analysis of benchmarking data}, Journal
  of Global Optimization 59 (2014), pp. 259--275.

\bibitem{dapm2002a}
K. Deb, A. Pratap, S. Agarwal, and T. Meyarivan, \emph{{A Fast and Elitist
  Multiobjective Genetic Algorithm: NSGA-II}}, IEEE Transactions on
  Evolutionary Computation 6 (2002), pp. 182--197.

\bibitem{DFSW2009}
B. Doerr, M. Fouz, M. Schmidt, and M. Wahlstr{\"o}m, \emph{{BBOB}:
  {N}elder-{M}ead with resize and halfruns}, in Rothlauf
  \cite{DBLP:conf/gecco/2009c}, 2009, pp. 2239--2246.

\bibitem{DWYV2018}
C. Doerr, H. Wang, F. Ye, S. van  Rijn, and T. B{\"a}ck, \emph{{IOHprofiler}: A
  benchmarking and profiling tool for iterative optimization heuristics}, ArXiv
  e-prints \href{https://arxiv.org/abs/1810.05281}{arXiv:1810.05281} (2018).

\bibitem{DM2002}
E.D. Dolan and J.J. Mor{\'e}, \emph{Benchmarking optimization software with
  performance profiles}, Mathematical programming 91 (2002), pp. 201--213.

\bibitem{DN2011}
J.J. Durillo and A.J. Nebro, \emph{{jMetal: a Java Framework for
  Multi-Objective Optimization}}, Advances in Engineering Software 42 (2011),
  pp. 760--771.

\bibitem{EFR1994}
B. Efron and R.J. Tibshirani, \emph{An introduction to the bootstrap}, CRC
  press, 1994.

\bibitem{MIH2006}
T.A. El-Mihoub, A.A. Hopgood, L. Nolle, and A. Battersby, \emph{Hybrid genetic
  algorithms: A review.}, Engineering Letters 13 (2006), pp. 124--137.

\bibitem{FIN2010}
S. Finck, N. Hansen, R. Ros, and A. Auger, \emph{Real-parameter black-box
  optimization benchmarking 2009: Presentation of the noiseless functions},
  Tech. {R}ep. 2009/20, Research Center PPE,  2009.

\bibitem{fletcher1970bfgs}
R. Fletcher, \emph{A new approach to variable metric algorithms}, Computer
  journal 13 (1970), pp. 317--322.

\bibitem{GKLS2003}
M. Gaviano, D. Kvasov, D. Lera, and Y.D. Sergeyev, \emph{Software for
  generation of classes of test functions with known local and global minima
  for global optimization}, ACM Transactions on Mathematical Software  (2003),
  pp. 469--480.

\bibitem{georges2018feature}
A. Georges, A. Gleixner, G. Gojic, R.L. Gottwald, D. Haley, G. Hendel, and B.
  Matejczyk, \emph{Feature-based algorithm selection for mixed integer
  programming}, Tech. {R}ep. 18-17, ZIB, Takustr. 7, 14195 Berlin,  2018.

\bibitem{goldfarb1970bfgs}
D. Goldfarb, \emph{A family of variable metric updates derived by variational
  means}, Mathematics of Computation 24 (1970), pp. 23--26.

\bibitem{GOU2016}
N. Gould and J. Scott, \emph{A note on performance profiles for benchmarking
  software}, ACM Transactions on Mathematical Software (TOMS) 43 (2016).

\bibitem{HAN2016perf}
N. Hansen, A. Auger, D. Brockhoff, D. Tu{\v s}ar, and T. Tu{\v s}ar,
  \emph{{COCO}: Performance assessment}, ArXiv e-prints
  \href{https://arxiv.org/abs/1605.03560}{arXiv:1605.03560} (2016).

\bibitem{HAN2009}
N. Hansen, A. Auger, S. Finck, and R. Ros, \emph{Real-parameter black-box
  optimization benchmarking 2009: Experimental setup}, Tech. {R}ep. RR-6828,
  INRIA,  2009. \urlprefix\url{http://hal.inria.fr/inria-00362649/en/}.

\bibitem{HAN2010ex}
N. Hansen, A. Auger, S. Finck, and R. Ros, \emph{Real-parameter black-box
  optimization benchmarking 2010: Experimental setup}, Tech. {R}ep. RR-7215,
  INRIA,  2010. \urlprefix\url{http://coco.gforge.inria.fr/bbob2010-downloads}.

\bibitem{HAN2016co}
N. Hansen, A. Auger, O. Mersmann, T. Tu{\v s}ar, and D. Brockhoff,
  \emph{{COCO}: A platform for comparing continuous optimizers in a black-box
  setting}, ArXiv e-prints
  \href{https://arxiv.org/abs/1603.08785}{arXiv:1603.08785} (2016).

\bibitem{HAN2009fun}
N. Hansen, S. Finck, R. Ros, and A. Auger, \emph{Real-parameter black-box
  optimization benchmarking 2009: Noiseless functions definitions}, Tech.
  {R}ep. RR-6829, INRIA,  2009.
  \urlprefix\url{http://hal.inria.fr/inria-00362633/en/}.

\bibitem{HAN2009noi}
N. Hansen, S. Finck, R. Ros, and A. Auger, \emph{Real-parameter black-box
  optimization benchmarking 2009: Noisy functions definitions}, Tech. {R}ep.
  RR-6869, INRIA,  2009. \urlprefix\url{http://hal.inria.fr/inria-00369466/en}.

\bibitem{ho2001a}
N. Hansen and A. Ostermeier, \emph{{Completely Derandomized Self-Adaptation in
  Evolution Strategies}}, Evolutionary Computation 9 (2001), pp. 159--195.

\bibitem{HAN2016ex}
N. Hansen, T. Tu{\v s}ar, O. Mersmann, A. Auger, and D. Brockhoff,
  \emph{{COCO}: The experimental procedure}, ArXiv e-prints
  \href{https://arxiv.org/abs/1603.08776}{arXiv:1603.08776} (2016).

\bibitem{HAN2010}
N. Hansen, A. Auger, R. Ros, S. Finck, and P. Po\v{s}\'{\i}k, \emph{Comparing
  results of 31 algorithms from the black-box optimization benchmarking
  {BBOB}-2009}, in \emph{GECCO '10: Proceedings of the 12th annual conference
  comp on Genetic and evolutionary computation}, New York, NY, USA. ACM, 2010,
  pp. 1689--1696.

\bibitem{HAR1999}
G. Harik and F. Lobo, \emph{{A parameter-less genetic algorithm}}, in
  \emph{Genetic and Evolutionary Computation Conference (GECCO~1999)}, Vol.~1.
  ACM, 1999, pp. 258--265.

\bibitem{HS1981}
W. Hock and K. Schittkowski, \emph{Test Examples for Nonlinear Programming
  Codes}, Lecture Notes in Economics and Mathematical Systems Vol. 187,
  Springer, 1981.

\bibitem{HMSW1953}
A. Hoffman, M. Mannos, D. Sokolowsky, and N. Wiegmann, \emph{Computational
  experience in solving linear programs}, Journal of the Society for Industrial
  and Applied Mathematics 1 (1953), pp. 17--33.

\bibitem{HOO1995}
J.N. Hooker, \emph{Testing heuristics: We have it all wrong}, Journal of
  heuristics 1 (1995), pp. 33--42.

\bibitem{HOO1998}
H. Hoos and T. St\"utzle, \emph{Evaluating {L}as {V}egas Algorithms---Pitfalls
  and Remedies}, in \emph{Proceedings of the Fourteenth Conference on
  Uncertainty in Artificial Intelligence (UAI-98)}. 1998, pp. 238--245.

\bibitem{HUN2007}
J.D. Hunter, \emph{Matplotlib: A 2d graphics environment}, Computing in science
  \& engineering 9 (2007), p.~90.

\bibitem{Huyer:1999}
W. Huyer and A. Neumaier, \emph{Global optimization by multilevel coordinate
  search}, J. of Global Optimization 14 (1999), pp. 331--355.

\bibitem{HN2009}
W. Huyer and A. Neumaier, \emph{Benchmarking of {MCS} on the noiseless function
  testbed}, Manuscript available at
  \url{http://www.mat.univie.ac.at/~neum/papers.html} (2009).

\bibitem{JOH2002}
D.S. Johnson, \emph{A theoretician’s guide to the experimental analysis of
  algorithms}, Data structures, near neighbor searches, and methodology: fifth
  and sixth DIMACS implementation challenges 59 (2002), pp. 215--250.

\bibitem{LOSH2012}
I. Loshchilov, M. Schoenauer, and M. Sebag, \emph{Black-Box Optimization
  Benchmarking of NIPOP-aCMA-ES and NBIPOP-aCMA-ES on the BBOB-2012 Noiseless
  Testbed}, in \emph{Companion Proceedings of the Genetic and Evolutionary
  Computation Conference (GECCO~2012)}, T. Soule, ed. ACM, 2012, pp. 269--276.

\bibitem{MER2015}
O. Mersmann, M. Preuss, H. Trautmann, B. Bischl, and C. Weihs, \emph{Analyzing
  the bbob results by means of benchmarking concepts}, Evolutionary computation
  23 (2015), pp. 161--185.

\bibitem{MGH1981}
J.J. Mor{\'e}, B.S. Garbow, and K.E. Hillstrom, \emph{Testing unconstrained
  optimization software}, ACM Transactions on Mathematical Software (TOMS) 7
  (1981), pp. 17--41.

\bibitem{MOR2009}
J.J. Mor{\'e} and S.M. Wild, \emph{Benchmarking derivative-free optimization
  algorithms}, SIAM Journal on Optimization 20 (2009), pp. 172--191.

\bibitem{nelder1965dsm}
J. Nelder and R. Mead, \emph{{The downhill simplex method}}, Computer Journal 7
  (1965), pp. 308--313.

\bibitem{nemirovski1995information}
A. Nemirovski, \emph{Information-based complexity of convex programming},
  Lecture Notes  (1995).

\bibitem{nesterov2018lectures}
Y. Nesterov, \emph{Lectures on convex optimization}, Vol. 137, Springer, 2018.

\bibitem{POW2006}
M.J.D. Powell, \emph{The {NEWUOA} software for unconstrained optimization
  without derivatives}, Large Scale Nonlinear Optimization  (2006), pp.
  255--297.

\bibitem{PRI1997}
K. Price, \emph{Differential evolution vs. the functions of the second {ICEO}},
  in \emph{Proceedings of the {IEEE} International Congress on Evolutionary
  Computation}, Piscataway, NJ, USA. IEEE, 1997, pp. 153--157.

\bibitem{robivc2005differential}
T. Robi{\v{c}} and B. Filipi{\v{c}}, \emph{Differential evolution for
  multiobjective optimization}, in \emph{Evolutionary Multi-Criterion
  Optimization (EMO 2005)}. Springer, 2005, pp. 520--533.

\bibitem{ROS2009}
R. Ros, \emph{Benchmarking the {NEWUOA} on the {BBOB}-2009 function testbed},
  in Rothlauf  \cite{DBLP:conf/gecco/2009c}, 2009, pp. 2421--2428.

\bibitem{DBLP:conf/gecco/2009c}
F. Rothlauf (ed.), \emph{Genetic and Evolutionary Computation Conference, GECCO
  2009, Proceedings, Montreal, Qu{\'e}bec, Canada, July 8-12, 2009, Companion
  Material}, ACM,  2009.

\bibitem{SCHI1987}
K. Schittkowski, \emph{More test examples for nonlinear programming codes},
  Lecture Nots in Economics and Mathematical Systems Vol. 282, Springer, 1987.

\bibitem{shanno1970bfgs}
D.F. Shanno, \emph{Conditioning of quasi-newton methods for function
  minimization}, Mathematics of Computation 24 (1970), pp. 647--656.

\bibitem{STE1946}
S.S. Stevens, \emph{et~al.}, \emph{On the theory of scales of measurement},
  Science  (1946), pp. 677--680.

\bibitem{TCZJ2017}
Y. Tian, R. Cheng, X. Zhang, and Y. Jin, \emph{Platemo: A matlab platform for
  evolutionary multi-objective optimization [educational forum]}, IEEE
  Computational Intelligence Magazine 12 (2017), pp. 73--87.

\bibitem{TUS2016}
T. Tu{\v{s}}ar and B. Filipi{\v{c}}, \emph{Performance of the {DEMO} Algorithm
  on the Bi-objective {BBOB} Test Suite}, in \emph{Companion Proceedings of the
  Genetic and Evolutionary Computation Conference (GECCO 2016)}. ACM, 2016, pp.
  1249--1256.

\bibitem{TUS2019}
T. Tu\v{s}ar, D. Brockhoff, and N. Hansen, \emph{Mixed-integer benchmark
  problems for single- and bi-objective optimization}, in \emph{Genetic and
  Evolutionary Computation Conference (GECCO 2019)}. ACM, 2019, pp. 718--726.

\bibitem{TUS2017}
T. Tu\v{s}ar, N. Hansen, and D. Brockhoff, \emph{{Anytime Benchmarking of
  Budget-Dependent Algorithms with the COCO Platform}}, in \emph{International
  Multiconference Information Society (IS 2017)}. ACM, 2017, pp. 47--50.

\bibitem{VAR2018}
K. Varelas, A. Auger, D. Brockhoff, N. Hansen, O.A. ElHara, Y. Semet, R.
  Kassab, and F. Barbaresco, \emph{A comparative study of large-scale variants
  of CMA-ES}, in \emph{International Conference on Parallel Problem Solving
  from Nature}. Springer, 2018, pp. 3--15.

\bibitem{VOL2019}
V. Volz, B. Naujoks, P. Kerschke, and T. Tu\v{s}ar, \emph{Single- and
  multi-objective game-benchmark for evolutionary algorithms}, in \emph{Genetic
  and Evolutionary Computation Conference (GECCO 2019)}. ACM, 2019, pp.
  647--655.

\bibitem{WHI1996}
D. Whitley, S. Rana, J. Dzubera, and K.E. Mathias, \emph{Evaluating
  evolutionary algorithms}, Artificial intelligence 85 (1996), pp. 245--276.

\bibitem{zhang2008rm}
Q. Zhang, A. Zhou, and Y. Jin, \emph{{RM-MEDA}: A regularity model-based
  multiobjective estimation of distribution algorithm}, IEEE Transactions on
  Evolutionary Computation 12 (2008), pp. 41--63.

\bibitem{zt1998b}
E. Zitzler and L. Thiele, \emph{{Multiobjective Optimization Using Evolutionary
  Algorithms - A Comparative Case Study}}, in \emph{Conference on Parallel
  Problem Solving from Nature (PPSN~V)}, LNCS Vol. 1498, Amsterdam. 1998, pp.
  292--301.

\end{thebibliography}
\begin{appendix}
\new{
\section{How We Chose Test Functions}
Our first and main test suite, {\ttfamily bbob}, took only known and relatively simple test functions as a starting point.
For choosing any specific test function to begin with, the function should model either
\begin{itemize}
\item a difficulty of continuous domain optimization known to be important in practice,
as for example multimodality or ill-conditioning or a ridge-like topology, or
\item a comprehensible difficulty that is likely to be relevant in practice at least sometimes, as for example ruggedness, or
\item a rather simple topology that every search algorithm should be able to deal with, like for example a linear slope.
\end{itemize}
\new{We also wanted the functions to be comprehensible, in order to facilitate interpretation,
and we required them to be scalable with the dimension.}
These functions were then (slightly) modified, mainly to make them less amenable to simple exploits.
We also paired up functions to understand the effect of a particular change on the algorithm performance. Furthermore, we chose to have only a two dozen of functions in order to be able to run repeated experiments over a range of dimensions in reasonable time and to incentivize manual checking of the results on each and every function.
}
\section{\del{Why BBOB is Not a Competition}\new{COCO Versus Competitive Testing}}
The main motivation behind our benchmarking effort is to be able to generate and assess \new{a comprehensive profile of the performance}
of \new{solvers} and understand why they perform well on some functions and not so well on others.
In order to understand the behaviour of a \new{solver}, it is of vital importance to understand the underlying function it has been run on as well as possible.
Therefore, functions can not be presented as black boxes to the scientific community \new{and should ideally be fully comprehensible}.
Additionally, if we want to be able to compare results with previously collected performance data,
the functions can not substantially change over time.

The incentives for a competition are, however, different. The main goal in a competition is to perform well rather than to understand algorithm behaviour. Hence, the competition designer should, in particular, take precautions to prevent exploits and overtuning.
The most effective way to prevent this is to \new{present} the functions as black boxes not only to the \new{solver} but also to the scientific community and to change them frequently.

We have taken precautions in the function definitions and in the experimental setup such that unintended exploits of trivial function properties are unlikely.
For example, the function optima as well as the optimal function values are not easily accessible in the API.

\new{%
However, intentional exploitation and thereby neglect of the prescribed experimental setup is not prevented this way.
}

\section{Details on the Used Biobjective Performance Measure}\label{app:mo}
When benchmarking multiobjective algorithms, \new{we must choose a}\del{a choice has to be made as to which} quality measure \new{to compare algorithms}\del{the algorithms are compared with}. In \COCO, \del{we (currently) use a normalized version of}\new{this quality measure is based on} the well-known hypervolume indicator, \new{$I_{\text{HV}}$,} also known as the $\mathcal{S}$-metric, and introduced as ``the size of the space covered'' \new{in} \cite{zt1998b}. In the following, we assume a generic search space $\Omega$, two objective functions, and that the ideal point 

\begin{equation*}
p^{\text{ideal}}=\left(\min_{x\in \Omega} f_1(x), \min_{x\in \Omega} f_2(x)\right)
\end{equation*}

\noindent and the nadir point 

\begin{equation*}
p^{\text{nadir}} = \left(
                    \max_{\begin{array}{c}\del{ x^*\in \Omega} x^* \text{ Pareto-optimal} \end{array}} f_1(x^*),
                    \max_{\begin{array}{c}\del{ x^*\in \Omega} x^* \text{ Pareto-optimal} \end{array}} f_2(x^*)
									 \right)
\end{equation*}

\noindent are known, which is the case for all current biobjective test suites in \COCO.
%\new{Recall that solution $x$ \emph{weakly dominates} solution $y$ (denoted by $x \preceq y$) when $f_i(x) \leq f_i(y)$, $i = 1, 2$. Solution $x$ \emph{dominates} solution $y$ (denoted by $x \prec y$) when $x \preceq y$ and $f(x) \neq f(y)$.}
\new{In order to be able to compare indicator values over different functions, dimensions, and instances\footnote{%
The algorithms themselves see the raw objective values which can span several orders of magnitudes, but they have access to the nadir point as the upper bound of the region of interest in objective space.}, the objective space is first normalized using the following transformation:
\begin{equation}\label{eq:norm}
f_i^{\text{N}} = \frac{f_i - p_i^{\text{ideal}}}{p_i^{\text{nadir}} - p_i^{\text{ideal}}} 
\end{equation}
for $i = 1, 2$. In the normalized objective space, the ideal and nadir points correspond to $(0, 0)$ and $(1, 1)$, respectively.
}

The quality indicator $I: 2^{\Omega}\rightarrow \R$ used in COCO to measure the quality of a biobjective algorithm $\mathcal{A}$ after $t$ function evaluations depends on the set $S\subseteq \Omega$ of all non-dominated solutions found by $\mathcal{A}$ within the first $t$ function evaluations and on whether the nadir point has been dominated in the first $t$ function evaluations\del{ or not} \new{(see also Equation \eqref{eq:indicator})}:
\begin{itemize}
	\item If the nadir point is dominated by an objective vector $f(s)$ with $s\in S$, the quality indicator $I(S)$ \new{equals} the hypervolume indicator $I_{\new{\text{HV}}}(S, r)$ of $S$ \new{on the normalized objective space with $(1, 1)$ as the reference point $r$}.
	\item If the nadir point has not been dominated in the first $t$ evaluations, the quality indicator $I(S)$ is the smallest distance of a \new{normalized} objective vector $f^{\new{\text{N}}}(s)$ with $s\in S$ to the \new{normalized} objective space \del{which is dominated by}\new{that dominates} the nadir point, \new{namely $[0, 1]^2$,} multiplied by $-1$ to allow for maximization of the quality indicator.
\end{itemize}

Note that both parts of the quality indicator align with a zero value at the border between the (objective) space dominating the reference point and the (objective) space not dominating it. The quality is positive in the former and negative in the latter case. More formally:%\niko{does that really help or give more information than the text? $I_{HV}$ is neither defined above nor here.}\tea{I think the equation helps and we should keep it, while I doubt a formal definition of $I_{HV}$ would be helpful in any way (and is probably safe to assume people that are interested in this part know what hypervolume is).}

\new{
\begin{equation}\label{eq:indicator}
I(S) = \left\{ 
\begin{array}{ll}
I_{\text{HV}}(S, \new{(1, 1)}) & 
\text{if } \exists s\in S: f_1(s) \leq p^{\text{nadir}}_1,  f_2(s) \leq p^{\text{nadir}}_2\\
-\min\limits_{\substack{s\in S\\z \in [0, 1]^2}}\text{dist}(f^\text{N}(s), z) & 
\text{otherwise}
\end{array}
\right..
\end{equation}
}

\del{In order to be able to compare indicator values over different functions, dimensions, and instances, before the calculation of the above indicator, the objective space is normalized. \todo{If this is done \emph{before} to compute $I_{HV}$, the nadir point should not be input to $I_{HV}$ but the input should be $1$ as value for the nadir!?}
To this end, the objective functions are scaled linearly\niko{isn't an \emph{affine} linear transformation needed? Why not write the transformation $(f_i - p^\text{ideal}_i) / p^\text{nadir}_i) $?} such that the ideal and nadir points are mapped to the all zeros and the all ones vector respectively.}

Finally, we record the number of function evaluations to reach certain\del{ absolute} target indicator values $$I_{\new{\text{HV}}}(S^*, \new{(1, 1)}) - \varepsilon,$$ where $I_{\new{\text{HV}}}(S^*, \new{(1, 1)})\in[0,1]$ is a reference hypervolume value, obtained as the hypervolume\del{value} of the best known Pareto set approximation $S^*$ \new{in the normalized objective space}, and\del{where} $\varepsilon$ is a target precision such as $10^{-5}$. Because $S^*$ is only an estimation of the true Pareto set, we also record runtimes for $\varepsilon = 0$ and for a few negative $\varepsilon$ values.

For a more detailed description and further illustrations, we refer the interested reader to \cite{BRO2016}.

\end{appendix}
\end{document}